\documentclass[twoside]{article}

\usepackage[accepted]{aistats2023}

\usepackage{algorithm}
\usepackage{algorithmic}

\usepackage{amsmath,amssymb,amsfonts,amstext,amsthm,mathrsfs}

\newtheorem{theorem}{Theorem}
\newtheorem{coro}{Corollary}
\newtheorem{lemma}{Lemma}

\newtheorem{assumption}{Assumption}

\usepackage{cleveref}

\usepackage{url}
\usepackage{enumitem}
\usepackage{makecell}
\usepackage{diagbox}
\usepackage{subfigure}
\usepackage{amsmath, bm}
\usepackage{dsfont}
\usepackage{amssymb}
\usepackage{multirow}
\usepackage{varwidth}
\usepackage{pifont}
\usepackage{makecell}
\usepackage{wrapfig}
\usepackage{multicol}
\usepackage{graphicx} 
\usepackage{comment}
\usepackage{color}
\usepackage{xcolor}
\usepackage{colortbl,booktabs}

\DeclareMathOperator*{\argmin}{arg\,min}
\usepackage[round]{natbib}

\begin{document}
\runningauthor{J. Yang$^\ast$, T. Chen$^\ast$, M. Zhu$^\ast$, F. He, D. Tao, Y. Liang, Z. Wang}

\twocolumn[
\aistatstitle{Learning to Generalize Provably in Learning to Optimize}

\aistatsauthor{Junjie Yang$^\ast$ \And Tianlong Chen$^\ast$ \And Mingkang Zhu$^\ast$ \And Fengxiang He \\ 
}

\aistatsaddress{ The Ohio State University \And  UT Austin \And UT Austin \And JD Explore Academy } 

\aistatsauthor{Dacheng Tao \And Yingbin Liang \And Zhangyang Wang }

\aistatsaddress{JD Explore Academy \And The Ohio State University \And UT Austin } 
]


\begin{abstract}
%
%
Learning to optimize (L2O) has gained increasing popularity, which automates the design of optimizers by data-driven approaches. However, current L2O methods often suffer from poor generalization performance in at least two folds: ($i$) applying the L2O-learned optimizer to unseen optimizees, in terms of lowering their loss function values \textit{(optimizer generalization, or ``generalizable learning of optimizers")}; and ($ii$) the test performance of an optimizee (itself as a machine learning model), trained by the optimizer, in terms of the accuracy over unseen data \textit{(optimizee generalization, or ``learning to generalize")}. While the optimizer generalization has been recently studied, the optimizee generalization (or learning to generalize) has not been rigorously studied in the L2O context, which is the aim of this paper. We first theoretically establish an implicit connection between the local entropy and the Hessian, and hence unify their roles in the handcrafted design of generalizable optimizers as equivalent metrics of the landscape flatness of loss functions. We then propose to incorporate these two metrics as \textit{flatness-aware} regularizers into the L2O framework in order to meta-train optimizers to learn to generalize, and theoretically show that such generalization ability can be learned during the L2O meta-training process and then transformed to the optimizee loss function. Extensive experiments consistently validate the effectiveness of our proposals with substantially improved generalization on multiple sophisticated L2O models and diverse optimizees. Our code is available at: \url{https://github.com/VITA-Group/Open-L2O/tree/main/Model_Free_L2O/L2O-Entropy}. 
\end{abstract}

\def\thefootnote{*}\footnotetext{The first three authors have made equal contributions.}
\section{Introduction}


One cornerstone of deep learning's success is the stochastic gradient-based optimization methods, such as SGD~\citep{10.2307/2236626}, Adam~\citep{kingma2014adam}, AdaGrad~\citep{JMLR:v12:duchi11a}, RProp~\citep{298623}, and RMSProp~\citep{Tieleman2012}. The performance of deep neural networks (DNNs) hinges on the choice of optimization methods and the corresponding parameter settings. Thus, intensive human labor is often required to empirically select the best optimization method and its parameters for each specific problem.

A promising data-driven approach, {\em learning to optimize (L2O)}, arises from the meta-learning community to alleviate this issue \citep{JMLR:v23:21-0308}. It aims to replace traditional optimization algorithms (i.e., {\bf optimizers}) tuned by humans, with optimizers parameterized by neural networks that can be trained to learn update rules from data. Existing works have demonstrated that such learned optimizers are able to decrease the objective function faster while tremendously reducing the required human labor. \citet{NIPS2016_fb875828} first proposed to parameterize the update rules using a long short-term memory (LSTM) network. The LSTM optimizer tries to simulate the behavior of iterative methods by unrolling. By aggregating a set of loss functions (i.e., {\bf optimizees}) to be optimized at each time step, it aims to minimize the overall loss along the optimization path. \citet{wichrowska2017learned} enlarged the optimizer model to a hierarchical recurrent neural network (RNN) to improve its capability on larger or unseen optimization problems. \citet{li2016learning} also proposed a reinforcement learning based L2O approach.



\begin{figure*}[t]
\centering
\subfigure[Optimizer Generalization]{\includegraphics[width=0.5\linewidth]{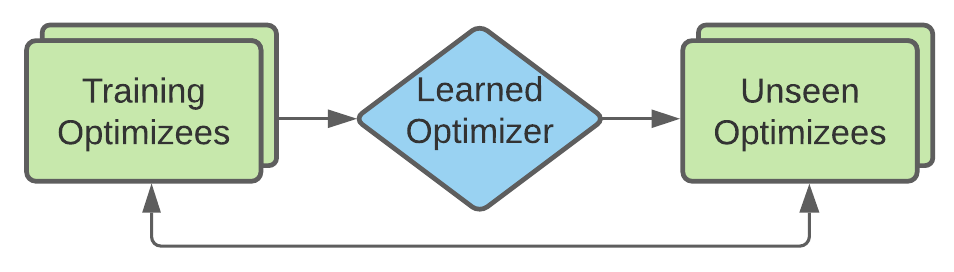}}
\subfigure[Optimizee Generalization]{\includegraphics[width=0.4\linewidth]{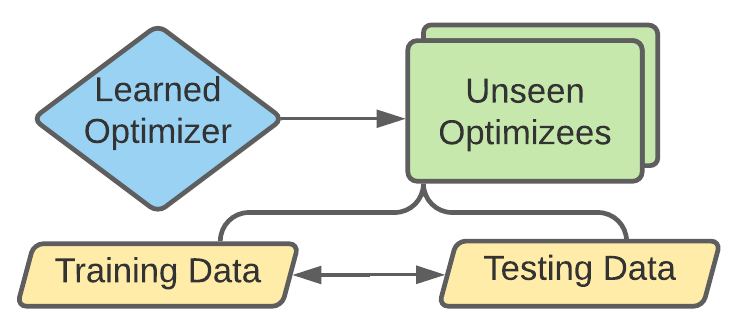}}
\caption{\small (a) Optimizer generalization characterizes the performance gap (training loss) between seen optimizees (during meta-training) and unseen optimizees (during meta-testing) by the same optimizer. (b) Optimizee generalization characterizes the performance gap between seen training data and unseen testing data, of the same unseen optimizee trained by an optimizer.}
\label{fig:ee_er_compare}
\end{figure*}

All existing L2O methods so far aim at the goal of ``learning to optimize", i.e., a meta-learned optimizer can minimize a given optimizee loss function successfully.
However, the generalization abilities, one of the core problems in machine learning, have not been explored thoroughly for current L2O methods. Specifically, there exist two different generalization concepts in the L2O context: {\bf optimizer generalization} (or ``generalizable learning of optimizers") and {\bf optimizee generalization} (or ``learning to generalize") (see \Cref{fig:ee_er_compare} for the difference between the two). {\em Optimizer generalization} characterizes how an optimizer trained by a certain set of given optimizees generalizes to unseen optimizees in terms of the unseen optimizee's training loss. On the other hand, {\em optimizee generalization} characterizes how an optimizee solution such as a classifier (trained by an optimizer) generalizes over the optimizee's unseen testing data. While the {\em optimizer generalization} has been recently studied in \citet{almeida2021generalizable}, the {\em optimizee generalization} has not been rigorously studied in the L2O context, which is the aim of this paper.


\subsection{Main Contributions}

This paper first examines the existing hand-crafted optimizer designs and
provides an unified understanding of the two core metrics used for facilitating the generalization ability. We then propose the ``learn to generalize" design, so that L2O can meta-train optimizers to have such generalization ability when they are applied to optimizees. 

In the traditional design of generalizable optimizers, the metrics of {\bf Hessian}~\citep{keskar2017large} and {\bf local entropy}~\citep{chaudhari2019entropy} are often adopted to directly design optimizee loss functions in order to achieve good generalization.
While {\em Hessian} directly measures the flatness of the loss landscape and facilitates the solution to a {\em flat basin}, the connection of {\em local entropy} to the loss geometry is rather implicit and has not been well understood. 
\begin{list}{$\bullet$}{\topsep=0.1ex \leftmargin=0.15in \rightmargin=0.1in \itemsep =0.0in}

\item The first contribution of the paper lies in establishing the implicit connection of local entropy to Hessian. Our theory explains that the existing Hessian and local entropy-based approaches are rooted in the same reason to improve the generalization performance. Specifically, we show theoretically that Hessian is upper bounded by a monotonically increasing function of the negative local entropy, and hence, large local entropy necessarily implies small Hessian. This explains that the Entropy-SGD algorithm~\citep{chaudhari2019entropy}, by minimizing the negative local entropy-based loss function, facilitates a model solution with small Hessian and hence a flat landscape. 


\end{list}

We then focus on the L2O problem and design a meta-training method for optimizers to ``learn to generalize".
\begin{list}{$\bullet$}{\topsep=0.1ex \leftmargin=0.15in \rightmargin=0.1in \itemsep =0.01in}


\item The second contribution of the paper lies in proposing to use the \textit{flatness-aware} regularizers based on Hessian and local entropy in the training of L2O optimizers. 
We show theoretically that such \textit{flatness-aware} regularizers in L2O can meta-train optimizers to have good generalization abilities, i.e., such trained optimizers will favor the convergence to flat landscape of the loss functions and hence enhance the generalization ability of their trained optimizees, even when the optimizees do not have a generalization-based design. Our theory shows that the generalization ability can be learned during the meta-training process and transformed to the optimizee.
\item We further provide comprehensive experiments over various tasks to demonstrate 
that our methods significantly improve the optimizee generalization ability of existing L2O methods, enabling them to outperform current state-of-the-art by a large margin. Our results also demonstrate that Hessian and local entropy yield very different practical performances. Local entropy is preferred when we adopt L2O to train large neural networks because it captures neighborhood landscape information which exhibits advantages in large neural networks. Instead, Hessian is preferred in small neural networks because it requires less time to compute.


\end{list}

\section{Related Work}
\paragraph{Learning to Optimize (L2O)} As a special case of learning to learn, L2O has been widely investigated in various machine learning problems~\citep{chen2017learning, cao2019learning, shen2021learning, li2020halo,chen2020automated, jiang2018learning,xiong2020improved, you2020l2, chen2020self, metz2020using, merchant2021learn2hop}. The first L2O framework dates back to~\citet{andrychowicz2016learning}, in which the gradients and update rules of optimizee are formulated as the input features and outputs for an RNN optimizer, respectively. Later on, \citet{li2016learning} proposes an alternative reinforcement learning framework for L2O, leveraging gradient history and objective values as observations and step vectors as actions. Recently, more advanced variants arise to power up the generalization ability of L2O. For example, ($i$) regularizers such as random scaling, objective convexifying~\citep{lv2017learning}, and Jacobian regularization~\citep{li2020halo}, ($ii$) enhanced L2O model such as hierarchical RNN architecture~\citep{wichrowska2017learned}, and ($iii$) improved training techniques such as curriculum learning and imitation learning~\citep{chen2020training}. Moreover, \citet{metz2021training} introduces randomly initialized optimizers to form a positive feedback loop for effective training. \citet{metz2019understanding} proposes a training scheme that dynamically weights two unbiased gradient estimators for a variational loss, and overcomes the strongly bias and exploding norm restrictions in L2O. Differently from the previous optimizee regularizer design, our proposed \textit{flatness-aware} regularizers are adopted to meta-train optimizers with good generalization abilities, even when the optimizees do not have generalization-based design.

\paragraph{Flatness on Generalization of Neural Networks}
Generalization analysis of neural networks has been widely studied by various methods, including VC-dimension \citep{bartlett2019nearly}, covering number \citep{bartlett2017spectrally}, stability \citep{hardt2016train, zhou2018generalization}, Rademacher complexity \citep{golowich2018size, ji2018minimax, ji2021understanding, arora2018stronger, arora2019fine}, etc. 
In particular, the landscape {\em flatness} has been known to be associated with better generalization. \citet{dinh2017sharp} showed theoretically that sharp minimum can also generalize well for deep neural networks, but such a result does not contradict the fact that flat minima generalize well, which has strong evidence \citep{he2019asymmetric, keskar2017large}.
On the empirical side, \citet{keskar2017large} and \citet{he2019asymmetric} showed that minima in wide valleys with small Hessian often generalize better than those in sharp basins with large Hessian. Further,  \citet{wilson2017marginal} and \citet{keskar2017improving} showed empirically that SGD favors better generalization solutions than Adam. On the theory side, \citet{zhou2020towards} showed that SGD is more unstable at sharp minima than Adam and \citet{zou2021understanding} explained that the inferior generalization performance of Adam is connected to nonconvex loss landscape. To improve the generalization performance, Entropy-SGD was introduced in \citet{chaudhari2019entropy} which was shown to outperform SGD in terms of the generalization error and the training time. Meanwhile, spectral norm regularization has been proposed in \citet{yoshida2017spectral} to improve the generalization ability of neural networks empirically. Further, \citet{foret2020sharpness} proposed the SAM method, which minimizes the loss value and the loss sharpness simultaneously. In this paper, we further explain the good generalization of Entropy-SGD by connecting the local entropy to the Hessian. We then further show that both Hessian and local entropy can serve as good regularizers to train L2O optimizers, which can yield optimizees with good generalization.
%

\paragraph{Learning to Generalize} Learning to generalize usually refers to domain generalization and domain adaptation~\citep{csurka2017domain}. Specifically, the goal of domain generalization is to learn a model which can generalize to unseen distributions and perform uniformly well across different data distributions~\citep{carlucci2019domain,dou2019domain,li2018learning}. Instead, this paper considers learning to generalize in the L2O context, which refers to the test performance of an optimizee and we call it \textit{optimizee generalization}. Another related but different generalization notation in L2O is \textit{optimizer generalization} or “generalizable learning of optimizers”, which was recently studied in ~\citet{almeida2021generalizable}. Such optimizer generalization characterizes the training loss of unseen optimizees when we apply the L2O-learned optimizer. 

\section{Local Entropy and Generalization of Entropy-SGD}
\label{sec:connection}
The local entropy was introduced in~\citet{chaudhari2019entropy} as the performance metric of the landscape of the loss function. Specifically, let $L(\theta)$ be a loss function, and define the {\bf local entropy} function of it as  
\[G(\theta;\gamma)  = \log \int_{\theta^\prime } \exp \left(-L(\theta^\prime)-\frac{\gamma}{2}\|\theta-\theta^\prime\|^2 \right) \text{d}\theta^\prime.\] Due to the exponential decay with respect to $\|\theta-\theta^\prime\|^2$, the integral mainly captures the value of the loss function $L(\theta^\prime)$ over the neighborhood of $\theta$. The local entropy has been applied in~\citet{chaudhari2019entropy} to design Entropy-SGD, and the authors of \citet{chaudhari2019entropy} demonstrated that Entropy-SGD enjoys better Lipschitz and smoothness conditions while favoring better generalization solutions empirically. However, it is still not well understood what type of geometry the local entropy measures and why it facilitates improving the generalization performance. 

Our following theorem establishes the implicit connection
between the local entropy and Hessian and thus explains that the local entropy also measures the flatness of the loss function. In this way, minimization of the local entropy yields models in the flat landscape and hence with good generalization performance.
\begin{theorem}[Connection between local entropy and Hessian]\label{thm:entropy}
Consider a non-negative convex loss function $L(\theta)$, where $\theta \in \mathbb{R}^p$, and suppose Assumption \ref{ass:Lip} in \Cref{subsec:assumption_1} holds. 
Then, 
\begin{align*}
    \|\nabla^2 L(\theta)\| \leq D^{-1}(-G(\theta;\gamma)),
\end{align*}
where $D^{-1}(x)$ is a monotonically increasing function, defined as the inverse function of $D(x)=\log(x+\gamma)+L(\theta) + (p-1)\log \gamma -mM-\frac{p}{2} \log(2\pi) - \frac{1}{2}\rho m^3- C(\gamma, p, m)$, $C(\gamma, p, m)=\log \int_{\theta^\prime} \exp \left(-\frac{\gamma}{2}\|\theta-\theta^\prime\|^2 \right) \text{d}\theta^\prime-\log \int_{\theta^\prime: \|\theta^\prime-\theta\|\leq m} \exp \left(-\frac{\gamma}{2}\|\theta-\theta^\prime\|^2 \right) \text{d}\theta^\prime$, $\theta\in\mathbb{R}^p$, $m$, $M$ and $\rho$ are constants.

\end{theorem}
It can be easily observed that the function $D(x)$ defined in \Cref{thm:entropy} is monotonically increasing w.r.t.\ $x$, as determined by the only $x$-dependent term $\log(x+\gamma)$, and hence its inverse $D^{-1}(x)$ is also monotonically increasing. 

\Cref{thm:entropy} shows that small negative local entropy $-G(\theta;\gamma)$ implies small Hessian $\|\nabla^2 L(\theta)\|$.  This explains that the Entropy-SGD algorithm, by minimizing a negative local entropy based loss function, facilitates a model solution with small Hessian and hence a flat landscape.


Note that~\citet{chaudhari2019entropy} proposed the Entropy-SGD method based on local entropy, and showed that the local entropy loss function enjoys better Lipschitz and smoothness conditions and hence favors better generalization solutions. As a comparison, we here establish the connection of local entropy to Hessian, and hence explain its better generalization via its landscape flatness.
\section{Learning to Generalize in L2O}\label{sec:method}
In this section, we provide basic notations and our design of ``learning to generalize" in L2O.

\subsection{Preliminary}\label{subsec:preli}
We define $l_{tr}(\theta; \xi)$ as the non-negative meta-training functions, where $\theta \in \mathbb{R}^p$ is the {\bf optimizee} parameter, and $\xi$ denotes training data samples. Suppose there are $N$ training data samples $\xi\in\{\xi_i, i=(1,\ldots,N)\}$. Then we define the empirical meta-training function and its corresponding population risk function as follows:
\begin{align}
\hat{L}_{tr}(\theta) = \frac{1}{N} \sum_{i=1}^{N} l_{tr}(\theta; \xi_i),  \enskip L_{tr}(\theta) =  \mathbb{E}_\xi l_{tr}(\theta;\xi).
\end{align}
An L2O algorithm aims to learn an \textit{update rule} for optimizee $\theta$ based on the meta-training function. An update rule can be expressed as $\theta^{t+1}_{tr}(\phi)=\theta_{tr}^t(\phi)+ m(z_{tr}^t; \phi)$, where $t=0,1,\ldots,T-1$ denotes the iteration index over one epoch, the variable $z$ captures the information (e.g., loss values, gradients) that we collect on the optimization path, and the {\bf optimizer} function $m(z;\phi)$ is parameterized by $\phi$ and captures how the update of the {\bf optimizee} parameter $\theta$ depends on the loss landscape information included in $z$. In order to find a desirable optimizer parameter $\phi$, L2O solves the following meta-training problem:
\begin{align}\label{eq:l2o}
 \nonumber \min_\phi & \{ \hat{L}_{tr}(\theta_{tr}^T(\phi))\} \\ 
 & \text{ where } \quad \theta_{tr}^{t+1}(\phi)=\theta_{tr}^t(\phi)+ m(z_{tr}^t;\phi).
\end{align}
A popular L2O meta-training algorithm applies the gradient descent method, which updates $\phi$ based on the gradient of the objective function $\hat{L}_{tr}(\theta_{tr}^T(\phi))$ with respect to $\phi$. As suggested by \cref{eq:l2o}, each update of $\phi$ requires $T$ iterations of the optimizee parameter $\theta_{tr}^0(\phi)$ to obtain $\theta_{tr}^T{(\phi)}$. 



\subsection{L2O Training via Flatness-aware Regularizers}
In order to train optimizers with generalization ability, 
we propose to incorporate \textit{flatness-aware} regularizers into L2O meta-training, so that such trained optimizers can learn to land the optimizee into a flat region, even if there is no generalization design for optimizee loss functions. We note that such an idea is fundamentally different from the existing handcrafted approaches, which directly design the optimizee loss function to feature flat landscapes. Rather, here we aim to let L2O auto-train optimizers to have generalization ability during meta-training, so that such optimizers will likely yield generalizable solutions when they are applied to optimizee loss functions, even when the optimizee function does not feature any flatness-aware design. We will show theoretically, such a ``learning to generalize" design will guarantee the transformation of generalization performance from meta-training to optimizee loss functions. Our result will also characterize the impact of the similarity between training and testing tasks as well as the difference between the training and testing losses on the generalization performance.




The first regularizer we introduce is based on the spectral norm of the {\em Hessian}, smaller values of which correspond to a flatter landscape.
Thus, the new L2O meta-training objective is given by:
\begin{align}\label{eq:hessian_obj}
\nonumber \min_\phi \{ \hat{L}_{tr}&(\theta_{tr}^T(\phi))+\lambda \|\nabla^2_\theta \hat{L}_{tr}(\theta^T_{tr}(\phi))\| \} \\ 
&\text{ where }\quad \theta^{t+1}_{tr}(\phi)=\theta^t_{tr}(\phi)+ m(z^t_{tr};\phi),
\end{align}
where $\lambda$ is the regularizer hyperparameter. Note that the Hessian regularizer is adopted for training the optimizer parameter $\phi$, and its impact on the update rule $m(z^t_{tr};\phi)$ is only through $\phi$, i.e., the information in $z^t$ does not include such regularization. Due to the computational intractability of directly penalizing $\nabla^2_\theta \hat{L}_{tr}(\theta^T_{tr}(\phi))$, we investigate three approximation variants in the implementation. The details can be found in \Cref{subsec:abla}. 


The second \textit{flatness-aware} regularizer we incorporate to L2O is based on the local entropy function defined in \Cref{sec:connection}. Specifically, consider the loss function  $\hat{L}_{tr}(\theta)$. Its local entropy is given by $\hat{G}_{tr}(\theta;\gamma)  = \log \int_{\theta^\prime } \exp \left(-\hat{L}_{tr}(\theta^\prime)-\frac{\gamma}{2}\|\theta-\theta^\prime\|^2 \right) \text{d}\theta^\prime$. Due to \Cref{thm:entropy}, the value of $\hat{G}_{tr}(\theta;\gamma)$ measures the flatness of the local area around $\theta$.
Thus, the L2O meta-training objective with the local entropy regularizer is given by:
\begin{align}\label{eq:entropy_obj}
\nonumber\min_\phi& \{ \hat{L}_{tr}(\theta_{tr}^T(\phi))-\lambda \hat{G}_{tr}(\theta^T_{tr}(\phi); \gamma) \}\\
&\text{ where }\quad \theta^{t+1}_{tr}(\phi)=\theta^t_{tr}(\phi)+ m(z^t_{tr};\phi).
\end{align}
In order to implement the gradient descent algorithm for meta-training, 
the gradient $-\nabla_\phi \hat{G}_{tr}(\theta_{tr}^T(\phi);\gamma)$ can be calculated by the entropy gradient $-\nabla_\theta \hat{G}_{tr}(\theta;\gamma)$ and the chain rule. In particular, as given in~\citet{chaudhari2019entropy}, the entropy gradient takes the following form
\begin{align}\label{eq:entropy_grad}
    -\nabla_\theta \hat{G}_{tr}(\theta;\gamma) = \gamma(\theta-\mathbb{E}[\theta^\prime;\xi]),
\end{align} 
where $\xi\in\{\xi_i, i=(1,\ldots,N)\}$ are training samples and the distribution of $\theta^\prime$ is given by $P(\theta^\prime;\theta, \gamma) \propto \exp \left[-\hat{L}_{tr}(\theta^\prime)-\frac{\gamma}{2}\|\theta-\theta^\prime\|^2\right]$. 





\subsection{Generalization Guarantee for Regularized L2O}
\label{subsec: genera_theory}
To analyze the optimizee generalization abilities, 
we first define $l_{ts}(\theta; \zeta)$ as the non-negative meta-testing function, where $\theta \in \mathbb{R}^p$ is the {\bf optimizee} parameter, and $\zeta$ denotes testing data samples. Suppose there are $M$ testing data samples $\zeta\in\{\zeta_i, i=(1,\ldots,M)\}$. Then we define the empirical meta-testing function and its corresponding population risk function as follows:
\begin{align}
 \hat{L}_{ts}(\theta) = \frac{1}{M} \sum_{j=1}^{M} l_{ts}(\theta; \zeta_j), \enskip
L_{ts}(\theta) =\mathbb{E}_\zeta l_{ts}(\theta;\zeta).
\end{align}
In meta-testing, we apply the output $\phi$ of meta-training and its corresponding optimizer to update the optimizee as $\theta_{ts}^{t+1}(\phi)=\theta_{ts}^{t}(\phi)+ m(z^t_{ts}; \phi)(t=0,1,\ldots,T-1)$. Note that we differentiate the optimizee updates in training and testing by subscripts $tr$ and $ts$, respectively.

Furthermore, we let $\phi^{\ast}$
be the optimal optimizer parameter for Hessian-regularized L2O, which can be written as
\begin{align}\label{eq:phistar}
\phi^{\ast} = \argmin_\phi \{ \hat{L}_{tr}(\theta_{tr}^T(\phi))+\lambda \|\nabla^2_\theta \hat{L}_{tr}(\theta_{tr}^T(\phi))\| \}.
\end{align}
Motivated by optimization theory, we note that the regularized optimization problem in \cref{eq:phistar} is equivalent to the following constrained optimization
\begin{align}\label{eq:hessian_obj_function}
\nonumber \min_\phi & \{\hat{L}_{tr}(\theta_{tr}^T(\phi))\} \quad \text{where} ~ \theta_{tr}^{t+1}(\phi)=\theta_{tr}^t(\phi)+ m(z^t_{tr};\phi) \\
&  \text{subject to } \|\nabla^2_\theta \hat{L}_{tr}(\theta_{tr}^T(\phi)) \| \leq B_{\text{Hessian}}(\lambda),
\end{align}
where $B_{\text{Hessian}}(\lambda)$ is the constraint bound on the Hessian determined by $\lambda$. Thus, the optimizer parameter $\phi^{\ast}$ learned by the Hessian-regularized L2O meta-training in \cref{eq:phistar} is also a solution to \cref{eq:hessian_obj_function}, i.e., its Hessian satisfies the constraint. Then we let $\theta^T_{ts}(\phi^{\ast})$ denote the optimizee parameters trained by optimizer $\phi^\ast$ in meta-testing, and $\theta_{ts}^{\ast}$ denote the optimal point of the population meta-testing function $L_{ts}(\cdot)$. We then characterize the generalization error as $L_{ts}(\theta^T_{ts}(\phi^{\ast}))-L_{ts}(\theta_{ts}^{\ast})$, which captures how well the optimizer $\phi^{\ast}$ performs on a testing task with respect to the best possible testing loss value. 

The following theorem characterizes the generalization performance of the optimizee trained with Hessian regularized optimizer as defined above. 
\begin{theorem}[Generalization Error of Hessian-Regularized L2O] \label{thm:hessian}
Suppose Assumptions \ref{ass:Lip}, \ref{ass:stronglyconvex},  \ref{ass:statistical_noise}, \ref{ass:stronglymorse} in \Cref{subsec:assumption_1} and \ref{subsec:assumption_2} hold. We let $N \geq \max \{4Cp \log N/\eta_{\ast}^2, Cp\log p\}$ where $C=C_0 \max\{c_h,1,\log(\frac{r\tau}{\delta})\}$, $\eta_{\ast}^2 = \min \{\frac{\epsilon^2}{\tau^2}, \frac{\eta^2}{\tau^4}, \frac{\eta^4}{\rho^2 \tau^2} \}$ and $C_0$ is a universal constant. Then, with probability at least $1-2\delta$, we have
\begin{align*}
\nonumber  L_{ts}&(\theta^T_{ts}(\phi^{\ast}))-L_{ts}(\theta_{ts}^{\ast}) \\
\leq & \textstyle \frac{1}{2} \Big(\Delta_T^{\ast} + \Delta_{\theta}^{\ast} + \mathcal{O}(w^{T-T^\prime})+ \mathcal{O}(\sqrt{\frac{C\log N}{N}})\Big)^2 \\ &\textstyle \Big(B_{\text{Hessian}}(\lambda) + \Delta_1^{\ast} +  \mathcal{O}(w^{T-T^\prime}) + \mathcal{O}(\sqrt{\frac{C\log N}{N}})\Big),
\end{align*}
where $\Delta_T^{\ast}=\| \theta^T_{ts}(\phi^{\ast}) - \theta_{tr}^T(\phi^{\ast}) \|$, $\Delta_\theta^{\ast}=\| \theta_{tr}^{\ast} - \theta_{ts}^{\ast}\|$, $\Delta_H^{\ast}=\| \nabla_\theta^2 L_{ts}(\theta_{ts}^\ast) - \nabla_\theta^2 L_{tr}(\theta_{tr}^\ast) \|$, $\Delta_1^\ast = \rho\Delta_T^{\ast} + \rho \Delta_{\theta}^{\ast} + \Delta_H^{\ast}$,  $w=\frac{L-\mu}{L+\mu}$, $T^\prime$ is the minimum gradient descent iterations for $\theta^{T^\prime}_{tr}(GD)$ to enter into the local basin of $\theta^T_{tr}(\phi^\ast)$ and GD refers to Gradient Descent.

\end{theorem}

\begin{figure*}[ht]
\centering
\includegraphics[width=1.0\linewidth]
{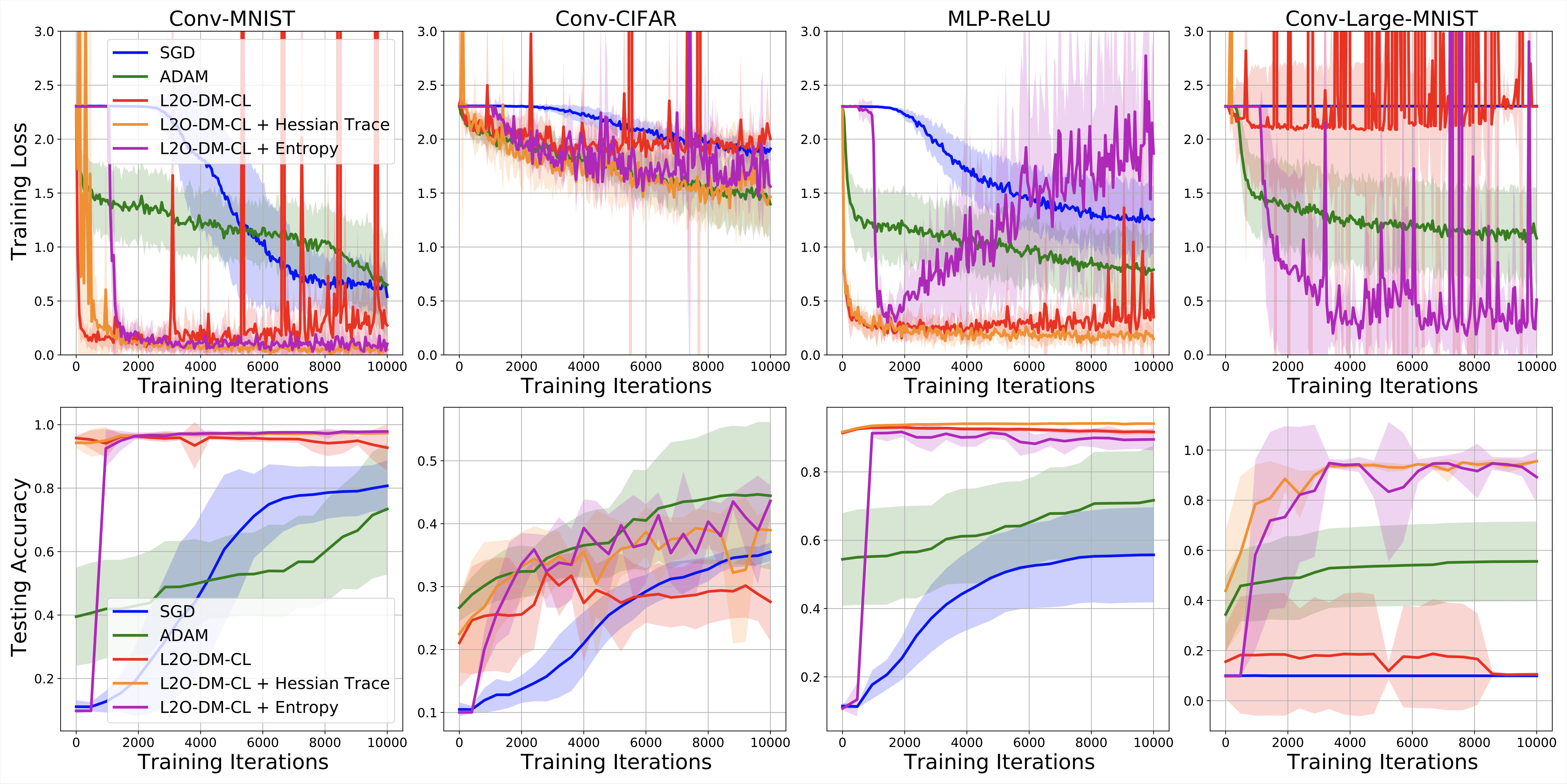}
\vspace{-6mm}
\caption{{\small Comparison of the training loss/testing accuracy of optimizees trained using analytical optimizers and L2O-DM-CL~\citep{chen2020training} with/without the proposed Hessian/Entropy regularization.}}
\vspace{-2mm}
\label{fig:hessian_dm}
\end{figure*}

\Cref{thm:hessian} characterizes the impact of the Hessian regularizer on the generalization error by the term 
$B_{\text{Hessian}}(\lambda)$. Clearly, by choosing the regularization hyperparameter $\lambda$, we control the value of $B_{\text{Hessian}}(\lambda)$ and further the generalization error. Specifically, larger $\lambda$ corresponds to smaller $B_{\text{Hessian}}(\lambda)$ and hence yields a smaller generalization error. This also explains that flatter landscape (i.e., smaller $B_{\text{Hessian}}(\lambda)$ on Hessian) yields better generalization performance (i.e., smaller generalization error).

The generalization error in \Cref{thm:hessian} also contains other terms which we explain as follows: (a) $\Delta_1^\ast = \rho\Delta_T^{\ast} + \rho \Delta_{\theta}^{\ast} + \Delta_H^{\ast}$ captures the similarities between the training and testing tasks: more similar tasks yield better generalization. These errors are owing to transformations of the generalization design in meta-training to optimizee training and testing; (b) $\mathcal{O}(w^{T-T^\prime})$ captures the exponential decay rate of the optimizee's iteration due to the strong convexity, and vanishes for large $T$; and (c) $\mathcal{O}(\sqrt{\frac{C\log N}{N}})$ arises due to the differences between the empirical and population loss functions, and vanishes as the sample size $N$ gets large.




We next analyze the generalization error of the Entropy regularizer on the optimizee generalization ability. Similarly to the Hessian regularizer, we let $\phi^{\ast}$
be the optimal optimizer parameter, which can be written as :
\begin{align}
\label{eq:phistarentropy}
\phi^{\ast} = \argmin_\phi \{\hat{L}_{tr}(\theta^T_{tr}(\phi))-\lambda \hat{G}_{tr}(\theta^T_{tr}(\phi); \gamma)\}.
\end{align}
Meanwhile, the regularized optimization problem in \cref{eq:phistarentropy} is equivalent to the following constrained optimization:
\begin{align}\label{eq:entropy_obj_function}
\nonumber \min_\phi & \{ \hat{L}_{tr}(\theta^T_{tr}(\phi))\} \quad \text{ where } \quad  \theta^{t+1}_{tr}=\theta^t_{tr}+ m(z^t_{tr};\phi) \\
&  \text{subject to } -\hat{G}_{tr}(\theta^T_{tr}(\phi); \gamma) \leq B_{\text{Entropy}}(\lambda),
\end{align}
where $B_{\text{Entropy}}(\lambda)$ is the constraint on the Entropy determined by $\lambda$. Thus, the optimizer $\phi^{\ast}$ learned by the Entropy-regularized L2O meta-training in \cref{eq:phistarentropy} is also a solution to \cref{eq:entropy_obj_function}, i.e., the local entropy satisfies the constraint.

\begin{coro}[Generalization Error of Entropy-Regularized L2O] \label{coro:entropy}
Suppose the same conditions of \Cref{thm:hessian} hold. Then the generalization error of L2O with Entropy regularizer takes the bound in \Cref{thm:hessian} with $B_{\text{Hessian}}(\lambda)$ being replaced by $D^{-1}(B_{\text{Entropy}}(\lambda))$.
\end{coro}

Corollary \ref{coro:entropy} shows that the bound $D^{-1}(B_{\text{Entropy}}(\lambda))$ serves the same role as the Hessian bound in the generalization performance. Thus, by controlling the hyperparameter $\lambda$ to be large enough in the L2O training, $B_{\text{Entropy}}(\lambda)$ as well as $D^{-1}(B_{\text{Entropy}}(\lambda))$ and Hessian can be controlled to be sufficiently small. In this way, the optimizee will be landed into a flat basin to enjoy better generalization.

\begin{figure*}[t]
\centering
\includegraphics[width=1.0\linewidth]
{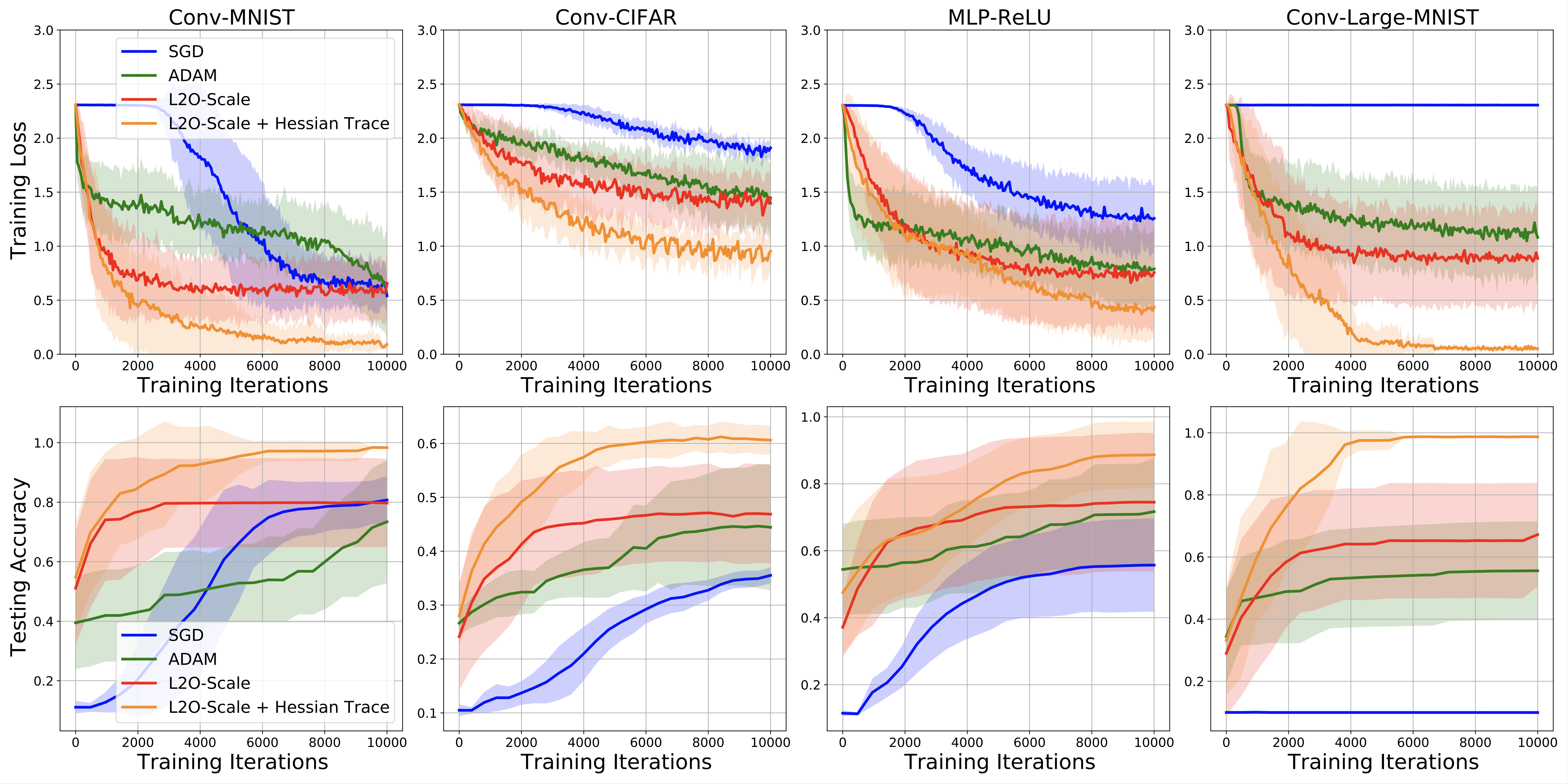}
\vspace{-6mm}
\caption{{\small Comparison of the training loss/testing accuracy of optimizees trained using analytical optimizers and L2O-Scale~\citep{wichrowska2017learned} with/without the proposed Hessian regularization.}}
 \vspace{-2mm}
\label{fig:hessian_scale}
\end{figure*}

\section{Experiments}
\label{experiment}

We consider two L2O algorithms: L2O-DM-CL\footnote{It is an enhanced version of the L2O-DM introduced by DeepMind \citet{andrychowicz2016learning}. We choose it as a stronger baseline with improved  generalization.}~\citep{chen2020training} and L2O-Scale~\citep{wichrowska2017learned}. For training L2O-Scale, we use a three-layer convolutional neural network (CNN) which has one fully-connected layer, and two convolutional layers with eight $3\times 3$ and $5\times 5$ kernels respectively. For training L2O-DM, we adopt the same meta training optimizee from~\citep{andrychowicz2016learning}, which is a simple Multi-Layer Perceptron (MLP) with one hidden layer of $20$ dimensions and the sigmoid activation. MNIST is used for all meta-training.


\begin{figure*}[ht]
\centering
\includegraphics[width=1.0\linewidth]
{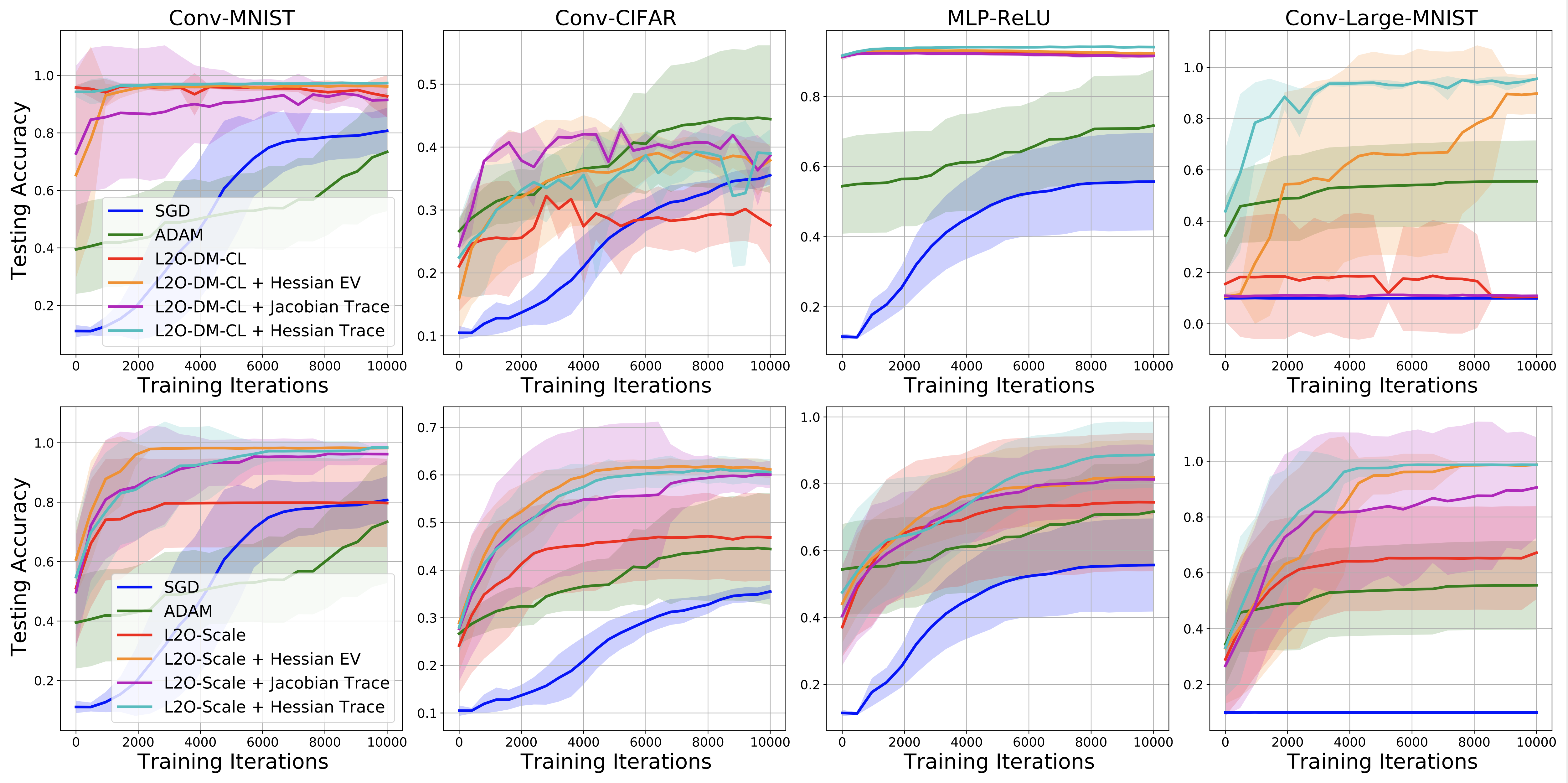}
\caption{{\small Comparison of the testing accuracy of optimizees trained using analytical optimizers and SOTA L2O with/without different Hessian regularization, \textit{Hessian EV}, \textit{Hessian Trace}, and \textit{Jacobian Trace}.}}
\label{fig:ablation_scale}
\end{figure*}

\vspace{-2mm}
\paragraph{Meta Testing Optimizees.} We select four distinct and representative meta testing optimizees from \citep{andrychowicz2016learning} and \citep{chen2020training} to evaluate the generalization ability of the learned optimizer. Specifically, \ding{172} \texttt{MLP-ReLU}: a single layer MLP with $20$ neurons and the ReLU activation function on MNIST. \ding{173} \texttt{Conv-MNIST}: a CNN has one fully-connected layer, and two convolutional layers with $16$ $3\times 3$ and $32$ $5\times 5$ kernels on MNIST. \ding{174} \texttt{Conv-Large-MNIST}: a large CNN has one fully-connected layer, and four convolutional layers with two $32$ $3\times 3$ and two $32$ $5\times 5$ kernels on MNIST. \ding{175} \texttt{Conv-CIFAR}: a CNN has one fully-connected layer, and two convolutional layers with $16$ $3\times 3$ and $32$ $5\times 5$ kernels on CIFAR-10~\citep{Krizhevsky09}. Optimizees \ding{172}, \ding{173}, and \ding{174} are for evaluating the generalization of L2O across network architectures. Then, \ding{175} evaluates the generalization of L2O across both network architectures and datasets.

\vspace{-3mm}
\paragraph{Training and Evaluation details.} 
During the meta-training stage of L2O, L2O-Scale is trained with $5$ epochs, where the number of each epoch's iteration is drawn from a heavy-tailed distribution~\citep{wichrowska2017learned}. L2O-DM-CL is trained with a curriculum schedule of training epochs and iterations, following the default setup in~\citet{chen2020training}. RMSprop with the learning rate $1 \times 10^{-6}$ is used to update L2Os. For the \{Hessian, Entropy\} regularization coefficients $\{\lambda_{\text{Hessian}},\lambda_{\text{Entropy}},\gamma\}$, we perform a grid search and choose \{$5\times10^{-5}$,-,-\}/\{$1\times10^{-10}$,$1\times10^{-6}$,$1\times10^{-4}$\} for L2O-Scale/L2O-DM-CL. 

In the meta-testing stage of L2O, we compare our methods with classical optimizers like SGD and Adam, and state-of-the-art (SOTA) L2Os such as L2O-Scale and L2O-DM-CL. Hyperparameters of both classical optimizers and L2O baselines are carefully tuned by the grid search and all other irrelevant variables are strictly controlled for a fair comparison. We run $10,000$ iterations for the meta-testing, and the corresponding training loss and test accuracy on all \textbf{unseen} optimizees are collected to evaluate the \textit{optimizer} and \textit{optimizee generalization}. We conduct \textbf{ten} independent replicates with different random seeds and all experiments are conducted on NVIDIA GeForce GTX 1080Ti GPUs.

\vspace{-2mm}
\subsection{Learning to Generalize with Hessian Regularization}
\label{subsec:exphessian}
\vspace{-2mm}
In this section, we conduct extensive evaluations of our proposed Hessian regularization on previous state-of-the-art L2O methods, i.e., L2O-Scale~\citep{wichrowska2017learned} and L2O-DM-CL~\citep{chen2020training}. Achieved training loss and testing accuracy are collected in Figure~\ref{fig:hessian_scale} and~\ref{fig:hessian_dm} which also include comparisons with representative analytical optimizers like SGD~\citep{ruder2016overview} and Adam~\citep{kingma2014adam}. Note that the training loss corresponds to $\hat{L}_{tr}$ and test accuracy corresponds to $L_{ts}$. Several consistent observations can be drawn from our results: 

\ding{182} Hessian Trace regularizer consistently enhances the generalization abilities of learned L2Os and trained optimizees. Specifically, L2Os with Hessian Trace enable fast training loss decay and much lower final loss on all four unseen meta-testing optimizees, showing the improved \textit{optimizer generalization} ability, which is great byproduct of our regularizer. Furthermore, all unseen optimizees trained by Hessian regularized L2Os enjoy substantial testing accuracy which boosts up to $31\%$, demonstrating the enhanced \textit{optimizee generalization} ability. Such impressive performance gains effective evidence of our proposal, which again suggests that Hessian regularization enables optimizers to learn to generalize.

\ding{183} Adopting vanilla L2O-DM-CL to train meta-testing optimizees (e.g., \texttt{Conv-MNIST} and \texttt{Conv-CIFAR}) suffers from instability as shown in Figure~\ref{fig:hessian_dm}, and it can be significantly mitigated by introducing our flatness-aware regularization. \texttt{Conv-Large-MNIST} is an exception, where the L2O-DM-CL fails to train this optimizee and ends up with random-guess accuracies, i.e., $10\%$. Although plugging Hessian Trace into L2O-DM-CL greatly improves its test accuracy from $10\%$ to $95\%+$, it still undergoes an unsatisfactory training loss. Reasons may lie in the rough model architecture and limited input features of L2O-DM-CL, coinciding with the findings in~\citep{chen2020training}. We will investigate this interesting phenomenon in the future. 

\ding{184} For advanced L2O-Scale, Hessian Trace regularization facilitates it to converge to significantly lower minima and to obtain considerable accuracy improvements. It enlarges the advantages of L2O methods compared to analytical optimizers, SGD and Adam, unleashing the power of parameterized optimizers.



\vspace{-2mm}
\subsection{Learning to Generalize with Entropy Regularization}
\vspace{-2mm}

We investigate the generalization improvements from the Entropy regularization. Generally, it boosts 
optimizee generalization
of L2O in most cases, as shown in Figure~\ref{fig:hessian_dm}. 

\vspace{-2mm}
\paragraph{Hessian v.s. Entropy Regularization.} We compare our two kinds of flatness-aware regularizers from both computational cost and performance benefits perspectives. 

\ding{182} In order to calculate the local entropy's gradient in~\cref{eq:entropy_grad}, it involves gradients from multiple unroll steps for the estimation~\citep{chaudhari2019entropy}, leading to extra memory and computing outlays. Compared to Hessian augmented L2O, it costs $\sim2.6$x memory and $\sim3$x running time for L2O-DM-CL experiments\footnote{We conduct entropy-related experiments on light-weight L2O-DM-CL rather heavy L2O-Scale models, since RTX TITAN with 24G memory is the largest GPU we can access and afford.}. Meanwhile, Hessian augmented L2O requires around 10\% more memory cost compared with vanilla L2O since we approximate Hessian in practice. However, the wall clock comparison included in \Cref{subsec:wall_clock} shows that these two regularizers share the same inference time which requires only $\sim1.5$x time than analytical optimizers. In comparison, another flatness-aware optimizer SAM~\citep{foret2020sharpness}, which incorporates loss landscape in the loss function, takes longer training time ($1.5$x $\sim$ $2$x SGD).

\ding{183} As for generalization gains, Entropy regularizer performs slightly better on \texttt{Conv-MNIST} and \texttt{Conv-CIFAR}, while behaves marginally worse on \texttt{MLP-ReLU} and \texttt{Conv-Large-MNIST} compared to Hessian regularizer. We would like to draw the reader's attention to \texttt{Conv-Large-MNIST}, in which Entropy regularized L2O-DM-CL is capable of decaying the training loss and finding a much lower minimum than Adam. Note that on this optimizee, both L2O-DM-CL and its Hessian variant can not decrease the training loss. The possible reason is that multi-layer convolutional neural networks without BN cannot be stably trained on MNIST. However, our L2O-DM-CL+Entropy is more stable in training and improves testing accuracy compared with L2O-DM-CL. This indicates that L2O-DM-CL + Entropy may also produce a more trainable loss surface for optimizees.

Based on the above experiments as well as those in \Cref{app:resnet20}, we observe that L2O+Entropy is preferred when we adopt L2O to train large neural networks, where L2O+Entropy yields better optimizer and optimizee generalization abilities. Meanwhile, L2O+Hessain optimizer requires less time per iteration to train and achieves lower training loss as well as higher test accuracy than L2O+Entropy in small MLPs. The possible reason is that Entropy takes account of the landscape over a large range of loss to measure the flatness, and can hence capture complex landscape information in large neural networks. On the other hand, the Hessian regularizer captures the flatness information only for the individual point, but in a more accurate manner, and thus is more suitable to smaller neural networks with a relatively simple landscape. We also compare our proposed methods with Entropy-SGD and SGD with Hessian regularization in \Cref{subsec:test_acc_comp} which demonstrates meta-training's advantages.


\vspace{-2mm}
\subsection{Ablation and Visualization}
\vspace{-2mm}
\label{subsec:abla}
In this section, we carefully examine the effect of Hessian regularization's different approximation variants, including
\ding{172} \textit{Hessian EV}:  the eigenvalue of largest module of Hessian matrix, computed by power iteration~\citep{yao2020pyhessian}; \ding{173} \textit{Hessian Trace}: the trace of Hessian matrix, calculated via Hutchinson method~\citep{yao2020pyhessian}; \ding{174} \textit{Jacobian Trace}: the trace of Hessian's Jacobian approximation $\nabla_\theta \hat{L}_{tr}(\theta^T_{tr}(\phi))^{\top}\nabla_\theta \hat{L}_{tr}(\theta^T_{tr}(\phi))$. Note that such Hessian approximation methods do not involve computing Hessian explicitly which helps to reduce the memory and computational cost and we adopt $10$ iterations for Hessian norms' approximation.
Results are presented in Figure~\ref{fig:ablation_scale}. We find that the Hessian Trace regularizer achieves the most stable and substantial performance boosts across all optimizees. Jacobian Trace performs the worst which is within expectation since it provides the roughest estimation of Hessian.

\section{Conclusion}
In this paper, we first establish an implicit connection between the local entropy and the Hessian. Then we propose \textit{flatness-aware} regularizers to incorporate these two metrics into the L2O framework for meta-training optimizers to learn to generalize. We further establish the theoretical guarantee to show that such generalization ability during L2O meta-training can be transformed to improve the optimizee's generalization over testing data.
Our empirical results validate the effectiveness of our proposal, taking a further step for L2O usage in real-world scenarios.

\section*{Acknowledgements}
The work of Y. Liang was supported in part by the U.S. National Science Foundation under the grant ECCS-2113860. The work of Z. Wang was supported in part by the U.S. National Science Foundation under the grant ECCS-2113904. 

\bibliography{L2o_aistats}

\clearpage

\appendix

\onecolumn

\noindent{\Large{\bf Supplementary Materials}}

\section{Additional Experimental Results} \label{append:add_exp}
\subsection{ResNet20 Experiments}\label{app:resnet20}
\begin{figure}[ht]
\centering
\includegraphics[width=0.75\linewidth]
{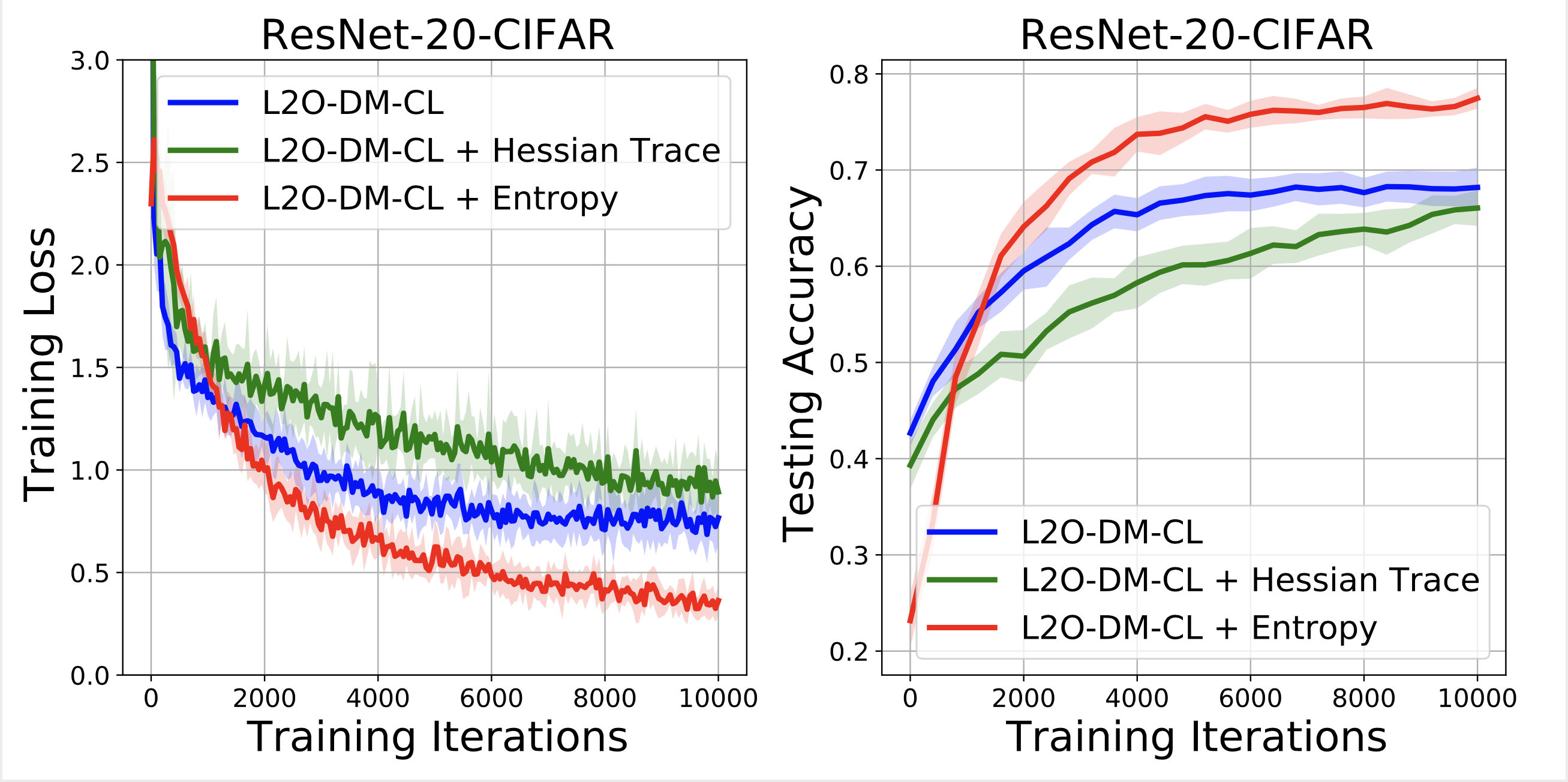}
\caption{{\small Comparison of the training loss/testing accuracy of ResNet-20 trained using L2O-DM-CL~\citep{chen2020training} with/without the proposed Hessian/Entropy regularization.}}
\label{fig:resnet}
\end{figure}

In this section, we evaluate the performance of our trained optimizers on larger neural networks ResNet-20 on CIFAR-10 dataset. The training loss and testing accuracy are plotted in \Cref{fig:resnet}. We can see that the Entropy regularizer is able to outperform other methods in both training loss and testing accuracy, demonstrating its generalization ability on large unseen models. Further note that although the Hessian regularizer may not be preferred in large neural networks, it does perform better than the Entropy regularizer in small networks as we have shown in \Cref{fig:hessian_dm}.

\subsection{Wall Clock Comparison between different algorithms}
\label{subsec:wall_clock}
We further conduct an optimizee training time comparison between our methods, analytical optimizers and L2O-DM-CL
in \Cref{table:time_result}. Note that L2O-DM-CL+Hessian and L2O-DM-CL+Entropy share the same time to train optimizee as L2O-DM-CL. From \Cref{table:time_result}, we can see that trained L2O-DM-CL requires only $\sim1.5$x time than analytical optimizers in terms of inference time, which is thus time efficient for practical usage. 

\begin{table}[ht]  
  \centering
  \caption{Empirical Time Cost Comparison per Iteration}
  \label{table:time_result}
\setlength{\tabcolsep}{2.5 mm}{
  \begin{tabular}{|l|ccc|}
  \hline
   Methods &  SGD & ADAM & L2O (\bf{L2O+Hessian, L2O+Entropy})  \\
 \hline
\hline
\multirow{1}{*}{Time (secs)}
       & 0.045 & 0.045 &  \bf{0.067}\\
\hline
  \end{tabular}
 }
\end{table}

\subsection{Accuracy Comparison between different algorithms}
\label{subsec:test_acc_comp}
We also compare the testing accuracy (\%) of our proposed methods with Entropy-SGD \citep{chaudhari2019entropy} and SGD with Hessian regularization. The Conv-MNIST results shown in \Cref{table:add1} are evaluated on L2O-DM-CL and the Conv-CIFAR results shown in \Cref{table:add2} are evaluated on L2O-Scale. We adopt the same experimental setting as in \Cref{experiment}  for the Conv-MNIST experiment. We also use same experimental setting for Conv-CIFAR except that the running epochs are set to 100 to investigate whether the performance of  trained optimizers would persist in long term.

\begin{table}[ht]  

  \centering
  \caption{Additional Testing Accuracy Comparison on Conv-MNIST}
  \label{table:add1}{
  \scalebox{0.95}[1]{
\setlength{\tabcolsep}{2.5 mm}{
  \begin{tabular}{|l|cccccc|}
  \hline
   Methods &  L2O & \bf{L2O+Hessian} & \bf{L2O+Entropy} & SGD & Entropy-SGD &  SGD+Hessian\\
 \hline
\hline
\multirow{1}{*}{Testing Accuracy}
       & 92.74 & \bf{97.34} &  \bf{97.87} &  80.73 & 97.54 &  95.37 \\
  \hline
  \end{tabular}
 }}}
\end{table}

\begin{table}[ht]  
  \centering
  \caption{Additional Testing Accuracy Comparison on Conv-CIFAR}
  \label{table:add2}{
  \scalebox{0.95}[1]{
\setlength{\tabcolsep}{2.5 mm}{
  \begin{tabular}{|l|cccc|}
  \hline
   Methods &  \bf{L2O+Hessian} & Entropy-SGD & SGD & SGD+Hessian\\
 \hline
\hline
\multirow{1}{*}{Testing Accuracy}
       & \bf{59.57} & 57.73 &  54.69 &   51.41\\
  \hline
  \end{tabular}
 }}}
\end{table}

From these comparisons, we can see that our proposed optimizers (L2O+Hessian, L2O+Entropy) achieve the best performance compared with regularized analytical optimizers. Specifically, in Conv-CIFAR setting as shown in \Cref{table:add2}, our algorithm L2O+Hessian outperforms SGD+Hessian and Entropy-SGD. In Conv-MNIST setting as shown in \Cref{table:add1}, the performances of top three algorithms, i.e. L2O+Entropy, Entropy-SGD and L2O+Hessian, are similar and much better than the performances of L2O and SGD+Hessian. Among the top three algorithms, the iteration running time for Entropy-SGD is 0.958 secs while L2O+Hessian and L2O+Entropy only take 0.067 secs as shown in \Cref{table:time_result}. Such wall clock comparison shows that L2O+Hessian and L2O+Entropy are more time efficient than Entropy-SGD while achieving the high accuracy, which are preferred for practical usage.

\subsection{Accuracy Comparison between different learning rates}
We further present the SGD and ADAM results within different learning rates as below:
\begin{table}[ht]  
  \centering
  \caption{Testing Accuracy Comparison of different SGD learning rates on Conv-MNIST}
  \label{table:add4}{
  \scalebox{0.95}[1]{
\setlength{\tabcolsep}{2.5 mm}{
  \begin{tabular}{|l|ccccc|}
  \hline
   SGD Learning Rate &  0.1   & 0.01   & 0.001 & 0.0001 & 0.00001\\
 \hline
\hline
\multirow{1}{*}{Testing Accuracy}
       & 9.81 & 78.01 & 80.84 & 11.09    & 10.82    
\\
  \hline
  \end{tabular}
 }}}
\end{table}

\begin{table}[!ht]  
  \centering
  \caption{Testing Accuracy Comparison of different ADAM learning rates on Conv-MNIST}
  \label{table:add5}{
  \scalebox{0.95}[1]{
\setlength{\tabcolsep}{2.5 mm}{
  \begin{tabular}{|l|ccccc|}
  \hline
   SGD Learning Rate &  0.1   & 0.01   & 0.001 & 0.0001 & 0.00001\\
 \hline
\hline
\multirow{1}{*}{Testing Accuracy}
       & 9.80 & 9.95 & 72.44 & 55.89 & 64.84    
\\
  \hline
  \end{tabular}
 }}}
\end{table}
Based on the \Cref{table:add4} and \cref{table:add5}, we know that both SGD and Adam are finetuned in terms of learning rates.

\section{Proof of \Cref{thm:entropy}} \label{appen: proofentropy}

\subsection{Assumptions} \label{subsec:assumption_1}

\begin{assumption}
\label{ass:Lip}
Lipschitz properties are assumed on functions $L_{tr}(\theta)$ and $L_{ts}(\theta)$.
\begin{list}{}{\topsep=0.ex \leftmargin=0.15in \rightmargin=0.in \itemsep =0.02in}
\item [a)] $L_{tr}(\theta)$ function is $M$-Lipschitz, i.e., for any $\theta_1$ and $\theta_2$, $\| L_{tr}(\theta_1) - L_{tr}(\theta_2) \| \leq M\|\theta_1-\theta_2\|$.
\item [b)] $\nabla_\theta L_{tr}(\theta)$ and $\nabla_\theta L_{ts}(\theta)$ are $L$-Lipschitz, i.e., for any $\theta_1$ and $\theta_2$, $\| \nabla_\theta L_{i}(\theta_1) - \nabla_\theta L_{i}(\theta_2) \| \leq L \|\theta_1-\theta_2\| (i=tr,ts)$.
\item [c)] $\nabla_\theta^2 L_{tr}(\theta)$ and $\nabla_\theta^2 L_{ts}(\theta)$ are $\rho$-Lipschitz, i.e.,  for any $\theta_1$ and $\theta_2$, $\|\nabla_\theta^2 L_{i}(\theta_1) - \nabla_\theta^2 L_{i}(\theta_2) \| \leq \rho\|\theta_1-\theta_2\| (i=tr,ts)$. This assumption also holds for stochastic $\nabla^2_\theta \hat{L}_{tr}(\theta)$ and $\nabla^2_\theta \hat{L}_{ts}(\theta)$. 
\end{list}
The above Lipschitz properties also hold for $L(\theta), \nabla_\theta L(\theta)$ and $\nabla^2_\theta L(\theta)$ in \Cref{thm:entropy}.
\end{assumption}

\subsection{Proof of Supporting Lemma}
\begin{lemma}
\label{lem:entropy}
Based on Assumption \ref{ass:Lip} and assuming that function $L(\theta)$ is non-negative and convex, in terms of entropy regularizer $G(\theta;\gamma)$, we have
\begin{align*}
    -G(\theta;\gamma) + mM + \frac{p}{2} \log(2\pi) + \frac{1}{2}\rho m^3+ C(\gamma, p, m) \geq&  \log(\det (\nabla^2L(\theta)+\gamma I))+L(\theta),
\end{align*}
where $m$ is a constant, $C(\gamma, p, m)=\log \int_{\theta^\prime} \exp \left(-\frac{\gamma}{2}\|\theta-\theta^\prime\|^2 \right) \text{d}\theta^\prime-\log \int_{\theta^\prime: \|\theta^\prime-\theta\|\leq m} \exp \left(-\frac{\gamma}{2}\|\theta-\theta^\prime\|^2 \right) \text{d}\theta^\prime$ and $\theta \in \mathbb{R}^p$.

\begin{proof}

We firstly split the integral area $\theta^\prime\in \mathbb{R}^p$ into two parts: $\{\theta^\prime:\|\theta^\prime-\theta\| \leq m\}$ and $\{\theta^\prime: \|\theta^\prime-\theta\|>m\}$. Based on the definition of $G(\theta;\gamma)$, we have
\begin{align*}
G&(\theta;\gamma) \\ 
= & \log \int_{\theta^\prime } \exp \left(-L(\theta^\prime)-\frac{\gamma}{2}\|\theta-\theta^\prime\|^2 \right) \text{d}\theta^\prime\\
     =& \log \int_{\theta^\prime: \|\theta^\prime-\theta\| \leq m} \exp \left(-L(\theta^\prime)-\frac{\gamma}{2}\|\theta-\theta^\prime\|^2 \right) \text{d}\theta^\prime \\
     &+ \int_{\theta^\prime: \|\theta^\prime-\theta\|> m} \exp \left(-L(\theta^\prime)-\frac{\gamma}{2}\|\theta-\theta^\prime\|^2 \right) \text{d}\theta^\prime \\
    \overset{(i)} \leq & \log \int_{\theta^\prime: \|\theta^\prime-\theta\| \leq m} \exp \left(-L(\theta^\prime) -\frac{\gamma}{2}\|\theta-\theta^\prime\|^2 \right) \text{d}\theta^\prime \\
    &+ \int_{\theta^\prime: \|\theta^\prime-\theta\|> m} \exp \left(-\frac{\gamma}{2}\|\theta-\theta^\prime\|^2 \right) \text{d}\theta^\prime \\
     \overset{(ii)}\leq& \log \int_{\theta^\prime: \|\theta^\prime-\theta\| \leq m} \exp \left(-L(\theta^\prime)-\frac{\gamma}{2}\|\theta-\theta^\prime\|^2 \right) \text{d}\theta^\prime + C(\gamma, p, m) \\
     \overset{(iii)}\leq& \log \int_{\theta^\prime: \|\theta^\prime-\theta\| \leq m} \exp \bigg(-L(\theta)-(\theta^\prime-\theta)^T\nabla L(\theta)-\frac{1}{2}(\theta^\prime-\theta)^T\nabla^2L(\theta^{\prime \prime})(\theta^\prime-\theta)\\
    &-\frac{\gamma}{2}\|\theta-\theta^\prime\|^2 \bigg) \text{d}\theta^\prime + C(\gamma, p, m), 
\end{align*}
where  $(i)$ follows from the fact that $L(\theta^\prime)$ is non-negative, $(ii)$ follows from the fact that $\theta \in \mathbb{R}^p$ and the definiton of $C(\gamma, p, m)$, $(iii)$ follows from Taylor expansion. Note that $\theta^{\prime\prime}$ satisfies $\|\theta^{\prime \prime}-\theta\| \leq \|\theta-\theta^\prime\|$ and $\|\theta^{\prime \prime}-\theta^\prime\| \leq \|\theta-\theta^\prime\|$.

Based on Assumption \ref{ass:Lip}, $-(\theta^\prime-\theta)^T\nabla L(\theta) \leq m\|\nabla L(\theta)\| \leq mM$. Then, we obtain
\begin{align*}
    G&(\theta;\gamma) \\
    \leq & -L(\theta)+m M+  \log \int_{\theta^\prime: \|\theta^\prime-\theta\| \leq m} \exp \bigg( -\frac{1}{2}(\theta^\prime-\theta)^T\nabla^2L(\theta^{\prime \prime})(\theta^\prime-\theta)-\frac{\gamma}{2}\|\theta-\theta^\prime\|^2 \bigg) \text{d}\theta^\prime + C(\gamma, p, m)\\
    =& -L(\theta)+mM+  \log \int_{\theta^\prime: \|\theta^\prime-\theta\| \leq m} \exp \bigg( -\frac{1}{2}(\theta^\prime-\theta)^T(\nabla^2L(\theta^{\prime \prime})+\gamma I)(\theta^\prime-\theta)\bigg) \text{d}\theta^\prime + C(\gamma, p, m) \\
    =& -L(\theta)+mM+C(\gamma, p, m) \\
    &+ \log \int_{\theta^\prime: \|\theta^\prime-\theta\| \leq m} \exp \bigg( -\frac{1}{2}(\theta^\prime-\theta)^T\Big(\nabla^2L(\theta^{\prime \prime})-\nabla^2L(\theta)+\nabla^2L(\theta)+\gamma I\Big)(\theta^\prime-\theta)\bigg) \text{d}\theta^\prime\\
    \overset{(i)}\leq & -L(\theta)+mM +C(\gamma, p, m) + \frac{1}{2}\rho m^3 \\
    &+\log \int_{\theta^\prime: \|\theta^\prime-\theta\| \leq m} \exp \bigg( -\frac{1}{2}(\theta^\prime-\theta)^T\left(\nabla^2L(\theta)+\gamma I\right)(\theta^\prime-\theta)\bigg) \text{d}\theta^\prime\\ 
    \overset{(ii)}\leq & -L(\theta)+mM + \frac{1}{2}\rho m^3+\log \int_{\theta^\prime} \exp \bigg( -\frac{1}{2}(\theta^\prime-\theta)^T\left(\nabla^2L(\theta)+\gamma I\right)(\theta^\prime-\theta)\bigg) \text{d}\theta^\prime + C(\gamma, p, m),
\end{align*}
where $(i)$ follows from Assumption \ref{ass:Lip} and the fact that $\|\theta^{\prime \prime}-\theta\|\leq \|\theta-\theta^\prime\|$ and $(ii)$ follows because $\exp ( -\frac{1}{2}(\theta^\prime-\theta)^T(\nabla^2 L(\theta)+\gamma I)(\theta^\prime-\theta)) \geq 0$.

Since $L(\theta)$ is convex, we have the fact that $(\nabla^2 L(\theta)+\gamma I)$ is a symmetric and positive-definite matrix. Hence, we obtain
\begin{align*}
      G&(\theta;\gamma) \leq -L(\theta) - \log(\det (\nabla^2L(\theta)+\gamma I)) + C(\gamma, p, m) +mM + \frac{1}{2}\rho m^3 + \frac{p}{2} \log(2\pi).
\end{align*}
We rearrange the terms and get
\begin{align*}
    -G(\theta; \gamma) \geq  L(\theta)  + \log(\det (\nabla^2L(\theta)+\gamma I))- C(\gamma, p, m)- mM - \frac{p}{2} \log(2\pi) - \frac{1}{2}\rho m^3.
\end{align*}

\end{proof}
\end{lemma}

\subsection{Proof of \Cref{thm:entropy}}
Based on \Cref{lem:entropy}, we have
\begin{align*}
    -G(\theta; \gamma)  + mM + \frac{p}{2} \log(2\pi) + \frac{1}{2}\rho m^3+ C(\gamma, p, m) \geq & \log(\det (\nabla^2L(\theta)+\gamma I))+L(\theta).
\end{align*}
Since $\nabla^2 L(\theta)+\gamma I$ is positive definite and $\lambda_i(\nabla^2L(\theta)+\gamma I)\geq \gamma$ for any $i=1,\ldots,p$. 
Then, based on the definition of Matrix norm
$\|\nabla^2 L(\theta)+\gamma I\| = \lambda_{\max} (\nabla^2L(\theta)+\gamma I)$,
we have
\begin{align*}
    \|\nabla^2L(\theta)+\gamma I\|^p \geq \det (\nabla^2L(\theta)+\gamma I)) \geq \gamma^{p-1}\|\nabla^2L(\theta)+\gamma I\|.
\end{align*}
Note that we use $\lambda_i(H)$ to denote the $i$-th eigenvalue of matrix $H$.
Then,
\begin{align*}
    \log(\det (\nabla^2L(\theta)+\gamma I))) \geq & (p-1)\log \gamma + \log \|\nabla^2L(\theta)+\gamma I\| \\
    = &(p-1)\log \gamma + \log(\|\nabla^2L(\theta)\|+\gamma).
\end{align*}
Then, we can obtain
\begin{align*}
    -G(\theta; \gamma)  + & mM + \frac{p}{2} \log(2\pi) + \frac{1}{2}\rho m^3+ C(\gamma, p, m) \geq L(\theta) + (p-1)\log \gamma \\
    &+ \log(\|\nabla^2L(\theta)\|+\gamma).
\end{align*}
Hence, we can get a new function $D(x)$ that
\begin{align*}
     \|\nabla^2L(\theta)\| \leq D^{-1}(-G(\theta; \gamma)),
\end{align*}
where $D(x)=L(\theta) + (p-1)\log \gamma -mM-\frac{p}{2} \log(2\pi) - \frac{1}{2}\rho m^3- C(\gamma, p, m)+\log(x+\gamma)$. Then, the proof is complete.


\section{Proof of \Cref{thm:hessian}}
\label{appen: proofhessian}

\subsection{Assumptions} \label{subsec:assumption_2}
We first define the local basin of $\theta$ with the radius $d$ as $D^d(\theta)=\{\theta^\prime: \|\theta-\theta^\prime\|_2 \leq d\}$. As have been observed widely in training a variety of machine learning objectives, the convergent point enters into a local neighborhood where the strong convexity (or similar properties such as gradient dominance condition, reguarity condition, etc) holds~\citep{du2018gradient, li2017convergence, zhou2018sgd, safran2016quality, milne2019piecewise}. We thus make the following assumption on the geometry of the meta-training function. 
\begin{assumption}\label{ass:stronglyconvex}
	We assume that there exist a a local basin $D^d(\theta_{tr}^T(\phi)) (d>0)$ of the convergence point $\theta_{tr}^T(\phi)$ that in such local basin, $L_{tr}(\theta)$ and $\hat{L}_{tr}(\theta)$ are $\mu$-strongly convex w.r.t. $\theta$. Futhermore, there exist a unique optimal point $\theta_{tr}^\ast$ of function $L_{tr}(\theta)$ and a optimal point $\hat{\theta}_{tr}^\ast$ of function $\hat{L}_{tr}(\theta)$ in local basin $D^d(\theta_{tr}^T(\phi^\ast))$. 


\end{assumption}

We further adopt the following assumptions introduced in \citet{mei2018landscape}, in order to guarantee the similarity between the landscape of the empirical and population objective functions.




\begin{assumption}\label{ass:statistical_noise}
	Similarly as in \cite{mei2018landscape}, we assume the loss gradient $\nabla l_{tr}(\theta;\xi)$ is $\tau^2$-sub-Gaussian, i.e., for any $\varrho \in \mathbb{R}^p$, and $\theta \in D^r(0)$ where $D^r(0) \equiv \{ \theta \in \mathbb{R}^p, \|\theta\|_2 \leq r \},$
	\begin{align*}
	\mathbb{E} \{ \exp (\langle \varrho, \nabla l_{tr}(\theta;\xi)-\mathbb{E}[\nabla l_{tr}(\theta; \xi)]\rangle )\} \leq \exp\left(\frac{\tau^2\| \varrho \|^2}{2}\right).
	\end{align*}
	Meanwhile, we assume the loss Hessian is $\tau^2$-sub-exponential, i.e., for any $\varrho \in D^1(0)$, and $\theta \in D^r(0)$,
	\begin{align*}
	\xi_{\varrho, \theta} \equiv \langle \varrho, \nabla^2 l_{tr}(\theta;\xi)\varrho \rangle, \quad
	\mathbb{E} \Big\{ \exp \left(\frac{1}{\tau^2}|\xi_{\varrho, \theta} - \mathbb{E} \xi_{\varrho, \theta}|  \right )  \Big\} \leq 2,
	\end{align*}
	 and there exists a constant $c_h$ such that $L\leq \tau^2 p^{c_h}, \rho \leq \tau^3 p^{c_h}$.
\end{assumption}

\begin{assumption}\label{ass:stronglymorse}
    We assume functions $L_{tr}(\theta)$ is $(\epsilon, \eta)$-strongly Morse in $D^r(0)$, i.e., if $\|\nabla L_{tr}(\theta)\|_2 > \epsilon$ for $\|\theta\|_2=r$ and, for any $\theta \in \mathbb{R}^p$, $\|\theta\|_2<r$,    the following holds:
	\begin{align*}
	\|\nabla L_{tr}(\theta)\|_2 \leq \epsilon \Rightarrow \min_{i \in [p]} |\lambda_i (\nabla^2 L_{tr}(\theta)) |\geq \eta,
	\end{align*}
	where $\lambda_i (\nabla^2 L_{tr}(\theta))$ denotes the $i$-th eigenvalue of $\nabla^2 L_{tr}(\theta)$. We further make the assumption that the local basins $D^d(\theta^T_{i}(\phi^\ast))(i=tr,ts)$ of convergence points $\theta^T_{i}(\phi^\ast) (i=tr,ts)$ are in $D^r(0)$.
\end{assumption}

\subsection{Proof of Supporting Lemmas}
\begin{lemma}[Restatement of Theorem 1(b) in \cite{mei2018landscape}] \label{lem: hessian}
	We assume $\theta^\ast$ corresponding to $\hat{\theta}^\ast$ in local basin. Based on Assumptions \ref{ass:Lip} and \ref{ass:statistical_noise}, there exists a universal constant $C_0$, and we let $C=C_0 \max \{c_h, \log(r\tau /\delta), 1 \}$. If $N \geq Cp\log p$, then we have
	\begin{align*}
	\sup_{\theta \in D^p(r)} \| \nabla^2 \hat{L}(\theta)-\nabla^2 L(\theta) \| \leq \tau^2\sqrt{\frac{Cp\log N}{N}},
	\end{align*}
	with probability at least $1-\delta$.
\end{lemma}

\begin{lemma}[Restatement of Theorem 2 in \cite{mei2018landscape}] \label{lem: theta}
Based on Assumptions \ref{ass:Lip}, \ref{ass:statistical_noise} and \ref{ass:stronglymorse}, we set $C$ as in \Cref{lem: hessian}, assume that $\theta^{\ast}$ is corresponding to $\hat{\theta}^\ast$, and let $N\geq 4Cp \log N/\eta_{\ast}^2$ where $\eta_{\ast}^2 = \min \{(\epsilon^2/\tau^2), (\eta^2/\tau^4), (\eta^4/(L^2\tau^2)) \}$.
Then, for each corresponding $\hat{\theta}^{\ast}$ and $\theta^{\ast}$, we have
\begin{align*}
\| \hat{\theta}^{\ast} - \theta^{\ast} \|_2 \leq \frac{2\tau}{\eta} \sqrt{\frac{Cp \log N}{N}},
\end{align*}
with probability at least $1-\delta$.
\end{lemma}

\begin{lemma}\label{lem:GD}
	Suppose Assumptions \ref{ass:Lip} and \ref{ass:stronglyconvex} hold. Then, we have
	\begin{align}
	\| \theta^T_{tr}(\phi^{\ast}) - \hat{\theta}^{\ast}_{tr} \| \leq \sqrt{\frac{L}{\mu}} \left( \frac{L-\mu}{L+\mu} \right)^{T-T^\prime} \|\theta^{T^\prime}_{tr}(GD) - \hat{\theta}^{\ast}_{tr}\|, 
	\end{align}
	where $T^\prime$ is the minimum that after $T^\prime$ gradient descent updates, the updated optimizee parameter $\theta^{T^\prime}_{tr}(GD)$ locates into the local basin of $\theta^T_{tr}(\phi^{\ast})$ and GD refers to Gradient Descent.
	\begin{proof}
	Since the local basin is $\mu$-strongly convex and $\hat{\theta}^{\ast}_{tr}$ is the optimal point of smooth function $\hat{L}_{tr}(\theta)$ in local basin. Then, we have
	\begin{align*}
	\hat{L}_{tr} (\theta^T_{tr}(\phi^{\ast})) - \hat{L}_{tr}(\hat{\theta}^{\ast}_{tr}) \geq \frac{\mu}{2} \|\theta^T_{tr}(\phi^{\ast})-\hat{\theta}^{\ast}_{tr}\|^2.
	\end{align*}
	Furthermore, we rearrange the terms and obtain
	\begin{align*}
	\|\theta^T_{tr}(\phi^{\ast})-\hat{\theta}^{\ast}_{tr}\|  \leq & \sqrt{\frac{2}{\mu}(\hat{L}_{tr} (\theta^T_{tr}(\phi^{\ast})) - \hat{L}_{tr}(\hat{\theta}^{\ast}_{tr})) } \\
	\overset{(i)}\leq & \sqrt{\frac{2}{\mu}(\hat{L}_{tr} (\theta^T_{tr}(GD)) - \hat{L}_{tr}(\hat{\theta}^{\ast}_{tr})) } \\
	\overset{(ii)}\leq & \sqrt{\frac{2}{\mu} \frac{L}{2} \|\theta^T_{tr}(GD) -\hat{\theta}^{\ast}_{tr}  \|^2} \\
	\leq & \sqrt{\frac{L}{\mu}} \|\theta^T_{tr}(GD) -\hat{\theta}^{\ast}_{tr}  \| \\
	\overset{(iii)}\leq &  \sqrt{\frac{L}{\mu}} \left( \frac{L-\mu}{L+\mu} \right)^{T-T^\prime} \|\theta^{T^\prime}_{tr}(GD) - \hat{\theta}^{\ast}_{tr}\|,
	\end{align*}
	where $(i)$ follows because $\phi^{\ast} = \argmin_\phi \hat{L}_{tr}(\theta^T_{tr}(\phi))$ and $\theta^T_{tr}(GD)$ locates in the local basin of $\hat{\theta}^{\ast}_{tr}$, $(ii)$ follows from Assumption \ref{ass:Lip} which implies $\hat{L}_{tr}(\theta)\leq \hat{L}_{tr}(\hat{\theta}_{tr}^\ast)+\langle\nabla_\theta \hat{L}_{tr}(\hat{\theta}_{tr}^\ast),\theta-\hat{\theta}_{tr}^\ast\rangle+\frac{L}{2}\|\theta-\hat{\theta}_{tr}^\ast\|^2$ and the fact that $\hat{\theta}^{\ast}_{tr} = \argmin_\theta \hat{L}_{tr}(\theta)$ which implies $\nabla_\theta \hat{L}_{tr}(\hat{\theta}_{tr}^\ast)=0$, and $(iii)$ follows if we set step size of GD as $\frac{2}{\mu+L}$.
	\end{proof}
\end{lemma}

\begin{lemma} \label{lem:hessianl}
	Based on Assumptions
	\ref{ass:Lip},
	\ref{ass:stronglyconvex},  \ref{ass:statistical_noise} and \ref{ass:stronglymorse}, we let  $N\geq \max \{Cp\log p, 4Cp \log N/\eta_{\ast}^2\}$ where $C=C_0 \max\{c_h,1,log(\frac{r\tau}{\delta})\}$, $\eta_{\ast}^2 = \min \{\frac{\epsilon^2}{\tau^2}, \frac{\eta^2}{\tau^4}, \frac{\eta^4}{\rho^2 \tau^2} \}$, $C_0$ is an universal constant. Then, with probability at least $1-2\delta$ we have
	
	\begin{align*}
	 \| \nabla_\theta^2 L_{ts}(\theta^\ast_{ts}) \| \leq &  \rho \left(\frac{2\tau}{\eta}\sqrt{\frac{Cp \log N}{N}} + \sqrt{\frac{L}{\mu}} \left( \frac{L-\mu}{L+\mu} \right)^{T-T^\prime} \|\theta^{T^\prime}_{tr}(GD) - \hat{\theta}^{\ast}_{tr}\| \right)\\
	 &+ \Delta_H^{\ast} +\tau^2\sqrt{\frac{Cp\log N}{N}}+ B_{\text{Hessian}}(\lambda),
	\end{align*}
	where $\Delta_H^{\ast}=\| \nabla_\theta^2 L_{ts}(\theta^\ast_{ts}) - \nabla_\theta^2 L_{tr}(\theta^\ast_{tr}) \|$, $T^\prime$ is defined in Lemma \ref{lem:GD} and GD refers to Gradient Descent.
	
	\begin{proof}
	Firstly, we bound $\| \nabla_\theta^2 L_{ts}(\theta^\ast_{ts}) \|  $ as following:
	\begin{align*}
	\| \nabla_\theta^2 L_{ts}(\theta^\ast_{ts}) \|  \leq &\| \nabla_\theta^2 L_{ts}(\theta^\ast_{ts}) - \nabla_\theta^2 L_{tr}(\theta^\ast_{tr}) \| + \| \nabla_\theta^2 L_{tr}(\theta^\ast_{tr}) - \nabla_\theta^2 \hat{L}_{tr}(\theta^\ast_{tr}) \| \\
	&+  \|\nabla_\theta^2 \hat{L}_{tr}(\theta^\ast_{tr})-\nabla_\theta^2 \hat{L}_{tr}(\hat{\theta}^\ast_{tr})\|+\|\nabla_\theta^2 \hat{L}_{tr}(
	\hat{\theta}^\ast_{tr}) - \nabla_\theta^2 \hat{L}_{tr}(\theta^T_{tr}(\phi^{\ast})) \| \\
	&+  \|\nabla_\theta^2 \hat{L}_{tr}(\theta^T_{tr}(\phi^{\ast}))\|,
	\end{align*}
	where $\theta^\ast_{tr}$ is corresponding to $\hat{\theta}^\ast_{tr}$ in the same local basin of $\theta^T_{tr}(\phi^{\ast})$.
	
	Based on the constrained problem formulation in \cref{eq:hessian_obj_function}, the optimal optimizer parameter $\phi^{\ast}$ is equivalent to the following: 
\begin{align*}
\phi^{\ast} = \argmin_\phi \hat{L}_{tr}(\theta^T_{tr}(\phi)) \text{ subject to }  \nabla^2_\theta \hat{L}_{tr}(\theta^T_{tr}(\phi)) \leq B_{\text{Hessian}}(\lambda).
\end{align*}
Thus, we obtain $ \|\nabla_\theta^2 \hat{L}_{tr}(\theta^T_{tr}(\phi^{\ast}))\| \leq B_{\text{Hessian}}(\lambda)$. Furthermore, if we let $N\geq Cp\log p$ where $C=C_0 \max\{c_h,1,log(\frac{r\tau}{\delta})\}$ and $C_0$ is an universal constant, based on Lemmas \ref{lem: hessian}, \ref{lem: theta} and \ref{lem:GD}, and Assumptions \ref{ass:Lip}, we have
	\begin{align*}
	 \| \nabla_\theta^2 L_{ts}(\theta^\ast_{ts}) \| \leq & \rho \left(\|\theta^\ast_{tr}-\hat{\theta}^{\ast}_{tr}\| + \sqrt{\frac{L}{\mu}} \left( \frac{L-\mu}{L+\mu} \right)^{T-T^\prime} \|\theta^{T^\prime}_{tr}(GD) - \hat{\theta}^{\ast}_{tr}\| \right)\\
	 & + \Delta_H^{\ast} + \tau^2\sqrt{\frac{Cp\log N}{N}} + B_{\text{Hessian}}(\lambda),
	\end{align*}
	with probability at least $1-\delta$.
	
	Furthermore, if we assume $N \geq \max \{4Cp \log N/\eta_{\ast}^2, Cp\log p\}$ where $\eta_{\ast}^2 = \min \{\frac{\epsilon^2}{\tau^2}, \frac{\eta^2}{\tau^4}, \frac{\eta^4}{\rho^2 \tau^2} \}$, based on Lemma  \ref{lem: theta}, we have
	
	\begin{align*}
	 \| \nabla_\theta^2 L_{ts}(\theta^\ast_{ts}) \| \leq&  \rho \left(\frac{2\tau}{\eta}\sqrt{\frac{Cp \log N}{N}}+\sqrt{\frac{L}{\mu}} \left( \frac{L-\mu}{L+\mu} \right)^{T-T^\prime} \|\theta^{T^\prime}_{tr}(GD) - \hat{\theta}^{\ast}_{tr}\| \right)\\
	 &+ B_{\text{Hessian}}(\lambda)+\Delta_H^{\ast} + \tau^2\sqrt{\frac{Cp\log N}{N}},
	\end{align*}
	with probability at least $1-2\delta$. 
		\end{proof}

\end{lemma}

\begin{lemma} \label{lem:theta_diff}
Based on Assumptions \ref{ass:stronglyconvex}, \ref{ass:Lip}, \ref{ass:statistical_noise}, and \ref{ass:stronglymorse}, we let $N\geq 4Cp \log N/\eta_{\ast}^2$ where $C$ and $\eta_{\ast}^2$ are defined in \Cref{lem: theta}. Then, with probability at least $1-\delta$, we have
\begin{align*}
\| \theta^T_{ts}(\phi^{\ast}) - \theta^\ast_{ts} \| \leq \Delta_T^{\ast} + \sqrt{\frac{L}{\mu}} \left( \frac{L-\mu}{L+\mu} \right)^{T-T^\prime} \|\theta^{T^\prime}_{tr}(GD) - \hat{\theta}^{\ast}_{tr}\|+ \frac{2\tau}{\eta} \sqrt{\frac{C p \log N}{N}} +\Delta_{\theta}^{\ast},
\end{align*}	
 where $\Delta_\theta^{\ast}=\| \theta^\ast_{tr} - \theta^\ast_{ts}\|$, $\Delta_T^{\ast}=\| \theta_{ts}^T(\phi^{\ast}) - \theta^T_{tr}(\phi^{\ast}) \|$ ,$T^\prime$ is defined in \Cref{lem:GD} and GD refers to Gradient Descent.
	
\begin{proof}
Based on triangle inequality, we obtain
	\begin{align*}
	\| &\theta^T_{ts}(\phi^{\ast}) - \theta^\ast_{ts} \| \\
	&\leq \|\theta^T_{ts}(\phi^{\ast})-\theta^T_{tr}(\phi^{\ast}) \| + \|\theta^T_{tr}(\phi^{\ast}) -\hat{\theta}^{\ast}_{tr} \| + \| \hat{\theta}^{\ast}_{tr} - \theta^\ast_{tr} \| + \|\theta^\ast_{tr}-\theta^\ast_{ts} \| \\
	&\overset{(i)} \leq \Delta_T^{\ast} + \|\theta^T_{tr}(\phi^\ast) - \theta^{\ast(1)}_N \|+ \| \hat{\theta}^{\ast}_{tr} - \theta^\ast_{tr} \| + \|\theta^\ast_{tr}-\theta^\ast_{ts} \| \\
	& \overset{(ii)} \leq \Delta_T^{\ast} + \sqrt{\frac{L}{\mu}} \left( \frac{L-\mu}{L+\mu} \right)^{T-T^\prime} \|\theta^{T^\prime}_{tr}(GD) - \hat{\theta}^{\ast}_{tr}\|+ \| \hat{\theta}^{\ast}_{tr} - \theta^\ast_{tr} \| +\Delta_{\theta}^{\ast},
	\end{align*}
	where $(i)$ follows from definition of $\Delta_T^\ast$, $(ii)$ follows from Lemma \ref{lem:GD} and definition of $\Delta_\theta^\ast$. Based on \Cref{lem: theta}, if we let $N\geq 4Cp \log N/\eta_{\ast}^2$. Then, with probability at least $1-\delta$, we have
	\begin{align*}
	\| \theta^T_{ts}(\phi^{\ast}) - \theta^\ast_{ts} \| \leq &  \Delta_T^{\ast} + \sqrt{\frac{L}{\mu}} \left( \frac{L-\mu}{L+\mu} \right)^{T-T^\prime} \|\theta^{T^\prime}_{tr}(GD) - \hat{\theta}^{\ast}_{tr}\|+ \frac{2\tau}{\eta} \sqrt{\frac{C p \log N}{N}} +\Delta_{\theta}^{\ast} \\
	= & \Delta_T^{\ast} + \Delta_{\theta}^{\ast} + \mathcal{O}(w^{T-T^\prime}) + \mathcal{O}(\sqrt{\frac{C\log N}{N}}), 
	\end{align*}
	where $w=\frac{L-\mu}{L+\mu}$.
\end{proof}
\end{lemma}

\subsection{Proof of \Cref{thm:hessian}}
Generalization loss is defined as below: 
\begin{align*}
L_{ts}&(\theta^T_{ts}(\phi^{\ast}))-L_{ts}(\theta^\ast_{ts}) \\
& \overset{(i)} = (\theta^T_{ts}(\phi^{\ast}) - \theta^\ast_{ts})^T \nabla_\theta L_{ts}(\theta^\ast_{ts}) + \frac{1}{2}(\theta^T_{ts}(\phi^{\ast}) - \theta^\ast_{ts})^T \nabla^2_\theta L_{ts}(\theta^{\prime})(\theta^T_{ts}(\phi^{\ast}) - \theta^\ast_{ts}) \\
& \overset{(ii)} = \frac{1}{2}(\theta^T_{ts}(\phi^{\ast}) - \theta^\ast_{ts})^T \nabla^2_\theta L_{ts}(\theta^{\prime})(\theta^T_{ts}(\phi^{\ast}) - \theta^\ast_{ts}) \\
& = \frac{1}{2}(\theta^T_{ts}(\phi^{\ast}) - \theta^\ast_{ts})^T (\nabla^2_\theta L_{ts}(\theta^{\prime})-\nabla^2_\theta L_{ts}(\theta^\ast_{ts})+\nabla^2_\theta L_{ts}(\theta^\ast_{ts}))(\theta^T_{ts}(\phi^{\ast}) - \theta^\ast_{ts}) \\
& \leq \frac{1}{2} \|\theta^T_{ts}(\phi^{\ast}) - \theta^\ast_{ts}\|^2(\| \nabla^2_\theta L_{ts}(\theta^{\prime})-\nabla^2_\theta L_{ts}(\theta^\ast_{ts})\| + \| \nabla^2_\theta L_{ts}(\theta^\ast_{ts})\|) \\
& \overset{(iii)}\leq \frac{1}{2} \rho \|\theta^T_{ts}(\phi^{\ast}) - \theta^\ast_{ts}\|^3 + \frac{1}{2} \|\nabla^2_\theta L_{ts}(\theta^\ast_{ts})\| \|\theta^T_{ts}(\phi^{\ast}) - \theta^\ast_{ts}\|^2 \\
& \leq \frac{1}{2} \|\theta^T_{ts}(\phi^{\ast})-\theta^\ast_{ts} \|^2 (\rho\|\theta^T_{ts}(\phi^{\ast})-\theta^\ast_{ts} \| + \|\nabla_\theta^2 L_{ts}(\theta^\ast_{ts}) \|),
\end{align*}
where $(i)$ follows from Taylor expansion and $\theta^{\prime}$ follows from the conditions that $\|\theta^{\prime} - \theta^T_{ts}(\phi^{\ast})\| \leq \|\theta^\ast_{ts}-\theta^T_{ts}(\phi^{\ast})\| $ and $\|\theta^{\prime} - \theta^\ast_{ts}\| \leq \|\theta^\ast_{ts}-\theta^T_{ts}(\phi^{\ast})\| $, $(ii)$ follows because $\nabla_\theta L_{ts}(\theta^\ast_{ts}) =0 $, and $(iii)$ follows from Assumption \ref{ass:Lip} and the fact that $\|\theta^{\prime} - \theta^\ast_{ts}\| \leq \|\theta^\ast_{ts}-\theta^T_{ts}(\phi^{\ast})\| $.

%

Based on Lemmas \ref{lem:hessianl} and \ref{lem:theta_diff}, if we let $N \geq \max \{4Cp \log N/\eta_{\ast}^2, Cp\log p\}$ where $C$ and $\eta_{\ast}^2$ are defined in \Cref{lem:hessianl}. Then, with probability at least $1-2\delta$, we have
\begin{align*}
 \rho\|\theta^T_{ts}&(\phi^{\ast})-\theta^\ast_{ts} \| + \|\nabla_\theta^2 L_{ts}(\theta^\ast_{ts}) \| \\
\leq & \rho \Big(\Delta_T^{\ast} + \sqrt{\frac{L}{\mu}} \left( \frac{L-\mu}{L+\mu} \right)^{T-T^\prime} \|\theta^{T^\prime}_{tr}(GD) - \hat{\theta}^{\ast}_{tr}\|+ \frac{2\tau}{\eta} \sqrt{\frac{C p \log N}{N}} +\Delta_\theta^{\ast}\Big)\\
&  + \tau^2\sqrt{\frac{Cp\log N}{N}} + \rho \left(\frac{2\tau}{\eta}\sqrt{\frac{Cp \log N}{N}} + \sqrt{\frac{L}{\mu}} \left( \frac{L-\mu}{L+\mu} \right)^{T-T^\prime} \|\theta^{T^\prime}_{tr}(GD) - \hat{\theta}^{\ast}_{tr}\|\right)+ \Delta_H^{\ast} + B(\lambda)  \\
=  & \rho \Delta_T^{\ast} + 2\rho \sqrt{\frac{L}{\mu}} \left( \frac{L-\mu}{L+\mu} \right)^{T-T^\prime} \|\theta^{T^\prime}_{tr}(GD) - \hat{\theta}^{\ast}_{tr}\| + \left(\frac{4\rho \tau}{\eta}+\tau^2\right) \sqrt{\frac{Cp\log N}{N}} + \rho \Delta_\theta^{\ast} + \Delta_H^{\ast} + B_{\text{Hessian}}(\lambda) \\
= & B_{\text{Hessian}}(\lambda) +\rho\Delta_T^{\ast} + \rho \Delta_{\theta}^{\ast} + \Delta_H^{\ast} +  \mathcal{O}(w^{T-T^\prime}) + \mathcal{O}(\sqrt{\frac{C\log N}{N}}), 
\end{align*}
where $w=\frac{L-\mu}{L+\mu}$, $\Delta_H^{\ast}=\| \nabla_\theta^2 L_{ts}(\theta^\ast_{ts}) - \nabla_\theta^2 L_{tr}(\theta^\ast_{tr}) \|$, $\Delta_\theta^{\ast}=\| \theta^\ast_{tr} - \theta^\ast_{ts}\|$, $\Delta_T^{\ast}=\| \theta_{ts}^T(\phi^{\ast}) - \theta^T_{tr}(\phi^{\ast}) \|$. 

Then, we have
\begin{align*}
    L_{ts}&(\theta^T_{ts}(\phi^{\ast}))-L_{ts}(\theta^\ast_{ts})\\
    &\leq \frac{1}{2} \|\theta^T_{ts}(\phi^{\ast})-\theta^\ast_{ts} \|^2 (\rho\|\theta^T_{ts}(\phi^{\ast})-\theta^\ast_{ts} \| + \|\nabla_\theta^2 L_{ts}(\theta^\ast_{ts})\|) \\
    & \leq \frac{1}{2} \left(\Delta_T^{\ast} + \Delta_{\theta}^{\ast} + \mathcal{O}(w^{T-T^\prime}) + \mathcal{O}(\sqrt{\frac{C\log N}{N}})\right)^2 \left(B_{\text{Hessian}}(\lambda) + \Delta_1^{\ast} +  \mathcal{O}(w^{T-T^\prime}) + \mathcal{O}(\sqrt{\frac{C\log N}{N}})\right)
\end{align*}
with probability at least $1-2\delta$ where $\Delta_1^\ast = \rho\Delta_T^{\ast} + \rho \Delta_{\theta}^{\ast} + \Delta_H^{\ast}$.  Then, the proof is complete.

\end{document}